\documentclass[twoside]{article}

\pdfoutput=1

\usepackage[accepted]{aistats2023}
\usepackage{booktabs}       %
\usepackage{nicefrac}       %
\usepackage{microtype}      %
\usepackage{wrapfig}
\usepackage{caption}
\usepackage{csquotes}
\usepackage{seqsplit}

\usepackage{tcolorbox}
\definecolor{cadmiumgreen}{rgb}{0.0, 0.42, 0.24}
\definecolor{oldmauve}{rgb}{0.4, 0.19, 0.28}
\definecolor{royalazure}{rgb}{0.0, 0.22, 0.66}
\definecolor{harvardcrimson}{rgb}{0.79, 0.0, 0.09}
\definecolor{lightmauve}{rgb}{0.86, 0.82, 1.0}
\definecolor{darkbrown}{rgb}{0.4, 0.26, 0.13}%
\usepackage[colorlinks = true,
            linkcolor = royalazure,
            urlcolor  = royalazure,
            citecolor = royalazure,
            anchorcolor = royalazure]{hyperref}
            
\usepackage{url}            %
\usepackage{floatrow} %
\usepackage{adjustbox}
\usepackage[noend]{algpseudocode}
\usepackage{amsfonts}
\usepackage{amsmath}
\usepackage{amssymb}
\usepackage{amsthm}
\usepackage{bm}
\usepackage{booktabs}
\usepackage{latexsym}
\usepackage{mathtools}
\usepackage{multirow}
\usepackage{paralist}
\usepackage{macros}
\usepackage{graphicx}
\usepackage{longtable}

\usepackage{amsmath,amsfonts,bm}

\def\1{\bm{1}}

\newcommand{\train}{\mathcal{D}}

\def\eps{{\epsilon}}

\def\vzero{{\bm{0}}}
\def\vone{{\bm{1}}}

\def\va{{\bm{a}}}

\def\vc{{\bm{c}}}

\def\vs{{\bm{s}}}

\def\vw{{\bm{w}}}
\def\vx{{\bm{x}}}

\def\mA{{\bm{A}}}

\def\mH{{\bm{H}}}
\def\mI{{\bm{I}}}

\def\mK{{\bm{K}}}

\def\mP{{\bm{P}}}

\def\mV{{\bm{V}}}

\def\mY{{\bm{Y}}}

\DeclareMathAlphabet{\mathsfit}{\encodingdefault}{\sfdefault}{m}{sl}
\SetMathAlphabet{\mathsfit}{bold}{\encodingdefault}{\sfdefault}{bx}{n}

\def\gF{{\mathcal{F}}}
\def\gG{{\mathcal{G}}}
\def\gH{{\mathcal{H}}}

\def\sP{{\mathbb{P}}}

\def\sR{{\mathbb{R}}}

\DeclareMathOperator{\ran}{ran}
\DeclareMathOperator{\spans}{span}

\newcommand{\ind}{\perp\!\!\!\!\perp} 

\newcommand{\E}{\mathbb{E}}

\newcommand{\Cov}{\mathrm{Cov}}

\DeclareMathOperator*{\argmin}{arg\,min}

\usepackage[round]{natbib}

\begin{document}

%

%
\runningauthor{Shaokui Wei, Jiayin Liu, Bing Li, Hongyuan Zha}

\twocolumn[

\aistatstitle{Mean Parity Fair Regression in RKHS}

\aistatsauthor{ Shaokui Wei \And Jiayin Liu}

\aistatsaddress{ Shenzhen Research Institute of Big Data \\ The Chinese University of Hong Kong, Shenzhen \And  School of Management and Economics\\ The Chinese University of  Hong Kong, Shenzhen} 

\aistatsauthor{Bing Li \And Hongyuan Zha }

\aistatsaddress{Department of Statistics \\ Pennsylvania State University \And School of Data Science \\ The Chinese University of Hong Kong, Shenzhen } 

]

\begin{abstract}
  We study the fair regression problem under the notion of Mean Parity (MP) fairness, which requires the conditional mean of the learned function output to be constant with respect to the sensitive attributes. We address this problem by leveraging reproducing kernel Hilbert space (RKHS) to construct the functional space whose members are guaranteed to satisfy the fairness constraints. The proposed functional space suggests a closed-form solution for the fair regression problem that is naturally compatible with multiple sensitive attributes. Furthermore, by formulating the fairness-accuracy tradeoff as a relaxed fair regression problem, we derive a corresponding regression function that can be implemented efficiently and provides interpretable tradeoffs. More importantly, under some mild assumptions, the proposed method can be applied to regression problems with a covariance-based notion of fairness.  Experimental results on benchmark datasets show the proposed methods achieve competitive and even superior performance compared with several state-of-the-art methods.
\end{abstract}

\section{INTRODUCTION}
    \label{sec::intro}
    	 As Machine Learning (ML) algorithms have been increasingly applied to solve real-world problems, such as employment \citep{kodiyan2019overview}, finance \citep{anshari2021financial}, and healthcare \citep{gupta2017advances}, the biases exhibited by  ML are attracting attention from both industry and academia. Algorithmic fairness has therefore emerged as a new frontier for ML, of which the critical challenge is to design algorithms satisfying fairness constraints, thus mitigating or eliminating the potential discrimination on the basis of legally protected (sensitive) attributes such as race or gender. In recent years, substantial efforts on notions and algorithms of fairness in ML have generally centered on classification problems \citep{agarwal2018reductions,calders2010three,  huang2019stable, jiang2020wasserstein, zafar2019fairness}, while the problems of fair regression have received much less attention.
  
    In this paper, we focus on the general regression problem in reproducing kernel Hilbert spaces (RKHS) and propose a novel approach for fair regression by constructing the space of functions that satisfy the constraints of fairness. Specifically, we consider the unfairness in the mean responses across different groups. Such unfairness exists broadly in many real-life problems including wage/payment gap \citep{oettinger1996statistical, barroso2021gender}, employment inequality \citep{pew2016views} and educational inequality \citep{darling1998unequal, baker2014school}. To mitigate such unfairness, we adopt the Mean Parity (MP) fairness, a notion of group fairness aiming to achieve "equality on average", i.e., the average response of a regression function to the different groups is the same.
 
    By establishing the connection between the covariance operator and MP fairness, we show that the MP-fair functional space can be characterized by a set of orthonormal bases and derive a closed-form solution that minimizes the mean squared error (MSE). Under some mild assumptions, the proposed method can also be applied to regression problems subject to fairness criterion that urges the outcome of the regression function to be uncorrelated with the sensitive attributes. In addition, the proposed method is naturally compatible with multiple sensitive attributes and can be extended to a broad range of loss functions for regression using optimization techniques, e.g., gradient descent.
	
    As it has been empirically observed that the fair model may suffer from a reduction in accuracy \citep{berk2017convex, tan2020learning}, we further generalize our method to consider the tradeoff between fairness and accuracy. By formulating the fairness-accuracy tradeoff as a relaxed fair regression problem, we derive a closed-form solution which is a simple combination of the optimal fair solution and the optimal least-squares solution, controlled by a single parameter. The proposed relaxed solution allows users to quantify and control the cost of fairness in terms of MSE and enjoys good interpretability. Finally, we evaluate our methods on three real datasets and one synthetic dataset. The experimental results demonstrate that our solution can eliminate the discrimination in train data and effectively enforce fairness in test data. Also, experiments on the fairness-accuracy tradeoff show that our method performs on par with other approaches and provides precise control over MSE and fairness levels.

    \paragraph{Paper organization.} The rest of the paper is organized as follows. Section \ref{sec::backg} introduces notations and the formulation of our problem. In Section  \ref{sec::fun}, we study the characterization of fair functional space in RKHS and provide a functional solution to the fair regression problem, after which we discuss the tradeoff between fairness and accuracy. Section \ref{sec::exp} presents some empirical evaluations of our methods. We discuss some related works in Section \ref{sec::rel} and end with some conclusions and future directions in Section \ref{sec::con}. The proofs, derivations, implementation details, and some additional experiments are left in Appendix.

\section{PRELIMINARIES}
    \label{sec::backg}
    \subsection{Notations}
We first introduce some important notations and a more comprehensive table of notations can be found in Appendix \ref{app::sec0}. Let $(\Omega, \gF, \sP)$ be a  probability space. We consider the random variables $X$ and $Y$ defined on measurable spaces  $(\Omega_X, \gF_X)$ and $(\Omega_Y, \gF_Y)$ where $\Omega_Y$ is a subset of $\sR$ and $\gF_Y$ is the Borel $\sigma$-filed on $\Omega_Y$. Let $\Omega_S=\{s^{(j)}\}_{j=1}^k$ be a finite set of $k$ elements from which a random variable $S$ takes values. We set $X$, $S$ and $Y$ to be the random variables for non-sensitive attributes, sensitive attributes and label/response respectively. In addition, we assume that $\sP(s)>0$ for all $s\in \Omega_S$.

Let $\kappa_X:\Omega_X\times \Omega_X\to \sR$ be a universal kernel and $\kappa_S: \Omega_S\times \Omega_S \to \sR$ be a discrete kernel. We use $\gH_X$ to represent the RKHS generated by $\kappa_X$ and denote its feature map by $\phi_X: \Omega_X\to \gH_{X}$, i.e., $\phi_X(x)=\kappa_{X}(\cdot, x)$. Similarly, let $\gH_S$ be the RKHS generated by $\kappa_S$ with feature map $\phi_{S}$. Let $\gH_{XS}$ be the RKHS generated by the kernel $\kappa_{XS}$ defined on $\Omega_{XS} \times \Omega_{XS}$ where $\Omega_{XS}=\Omega_{X}\times \Omega_{S}$. Then, each member of $\gH_{XS}$ is a function $g(x, s)$ where $x\in \Omega_X, s\in \Omega_S$, and we denote the feature map of $\gH_{XS}$ by $\phi_{XS}$. By the reproducing property of $\gH_{XS}$, evaluating a function $g\in \gH_{XS}$ at $(x,s)$ can be written as
\begin{equation*}
	g(x,s)=\left\langle \phi_{XS}(x,s), g\right\rangle_{\gH_{XS}},
\end{equation*}
where $\langle \cdot,\cdot \rangle_{\gH_{XS}}$ is the inner product in $\gH_{XS}$.
For a space $M\subseteq \gH_{XS}$, we denote its orthogonal complement in $\gH_{XS}$ by $M^\perp$ such that $\gH_{XS}=M\oplus M^\perp$. Moreover, let $\ind $  represent the independence between random variables.

\subsection{Notions of fairness}
Our goal is to find the optimal fair regression function in $\gH_{XS}$ that minimizes the mean squared error while maintaining fairness.  For a function $g\in \gH_{XS}$, we consider the Mean Parity \footnote{also known as Mean Difference \citep{calders2013controlling, vzliobaite2017measuring}, Mean Distance \citep{komiyama2017two} or  Discrimination Score \citep{calders2010three, pmlr-v28-zemel13, raff2018fair}.} fairness, as defined below:

\begin{definition}[Mean Parity]
	\label{def::MP}
	The subset $\gG_{MP}$ of $\gH_{XS}$ defined by 
	\begin{equation*}
		\label{eq::MP}
		\gG_{MP}=\{g\in \gH_{XS}: \E\left[g(X, S)|S\right]=\E\left[g(X, S)\right]\}
	\end{equation*}
	is called the Mean Parity fair (MP-fair) class of functions.
\end{definition}
The above definition says that a function $g$ is MP fair if the expectation of $g(X, S)$ conditioning on $S$ is constant across all sensitive groups.  

Besides MP-fairness, there are several other ways of defining fairness. Here, we highlight the connection and distinction between MP fairness and the other two notions of fairness. 

\paragraph{Demographic Parity (DP) fairness.} A popular requirement for fairness is $g(X, S) \ind S$, i.e., the distribution of $g(X,S)$ conditioning on $S$ is the same, and the class of such functions is called Demographic Parity fair class \citep{feldman2015certifying}.

To establish the relationship between MP fairness and DP fairness, we provide the following proposition:
\begin{proposition}
	\label{pro::mp}
	Assume that the DP disparity (DPD) and the MP disparity (MPD) of function $g$ are measured by
	\begin{align*}
		\text{DPD}(g)&=\sum_{s\in \Omega_S}\mathcal{W}_1(g(X,S)|S=s,g(X,S))\\
		\text{MPD}(g)&=\sum_{s\in \Omega_S}|\E(g(X,S)|S=s)-\E(g(X,S))|
	\end{align*}
	where $\mathcal{W}_1$ is the 1-Wasserstein distance \citep{frohmader20211} .
	
	Then, 
	$$\text{MPD}(g)\leq \text{DPD}(g).$$ 
\end{proposition}
Therefore, MP disparity is the lower bound of DP disparity and achieving MP fairness is necessary to achieve DP fairness. Moreover, for binary classification problem with a binary $S$, MP fairness is equivalent to DP fairness.  

\paragraph{Covariance based (CB) fairness.} Another widely used condition for fairness is the Covariance based (CB) fairness \citep{komiyama2018nonconvex,mary2019fairness, perez2017fair, scutari2021achieving}, which requires the output of $g$ to be uncorrelated with the sensitive attribute, i.e., $\Cov(g(X,S),S)=0$. 

By the definition of covariance, we can conclude that MP fairness implies that $g(X,S)$ is uncorrelated with $S$. Thus, an MP-fair regression function is always CB-fair. Moreover, MP fairness is equivalent to CB fairness under some assumptions, which will be discussed later.

\subsection{Problem formulation}
Now, we introduce the formulation for the MP-fair regression problem. Consider the general regression model
\begin{equation*}
	Y=g(X, S)+\eps,
\end{equation*}
where $g\in \gH_{XS}$ and $X, S$ are independent of the centered random noise $\eps\in \sR$ and $\E(Y^2)\leq \infty$.

Then, we focus on the least-squares MP-fair regression task formulated as a constrained optimization problem
\begin{equation}
	\label{eq::obj}
	\begin{aligned}
	\min_{g} \quad & \E\left(Y-g(X, S)\right)^2\\
	\mathrm{s.t. } \quad & g\in \gG_{MP}.
	\end{aligned}
 \end{equation}

\section{FAIR REGRESSION UNDER MEAN PARITY}
	\label{sec::fun}
	
In this section, we discuss how to solve Problem \ref{eq::obj}. To do so, we first develop a theory to characterize the MP-fair class $\gG_{MP}$ within $\gH_{XS}$. After that, we derive a closed-form solution by employing a projection operator $P$ from $\gH_{XS}$ onto $\gG_{MP}$ and introduce a formulation to control the fairness-accuracy tradeoff. At last, we discuss the performance guarantees of the derived solution and how to solve the MP-fair regression problem with other loss functions.
\subsection{Characterization of  MP-fair function space}
We begin by introducing some concepts. For Hilbert spaces $\gH_1$, $\gH_2$ and a linear operator $A:\gH_1\to \gH_2$, we define the kernel of $A$ by $\ker(A)=\{f\in \gH_1: Af=0_{\gH_2}\}$ where $0_{\gH_2}$ is the zero function in $\gH_2$.  Let $\ran(A)$ represent the set $\{Af: f\in \gH_1\}$, which is the range of $A$. Let $\mu_{XS}$ be the kernel mean embedding of $(X, S)$ in $\gH_{XS}$, which is defined as $\mu_{XS}= \E\left[\kappa_{XS}\left(\cdot, (X, S)\right)\right]$. Similarly, let $\mu_S = \E\left[
\kappa_S(\cdot, S)\right]$ be the kernel mean embedding of $S$  in $\gH_S$. Then, we define the covariance operator between $S$ and $(X, S)$ as 
\begin{equation*}
	\Sigma_{S(XS)}=\E\left[(\phi_{S}(S)-\mu_S)\otimes (\phi_{XS}(X, S)-\mu_{XS})\right],
\end{equation*}
where $\otimes $ represents the outer product in RKHS.

To characterize $\gG_{MP}$, we present the following assumption.
\begin{assumption}
	\label{ass:1}
	Assume that the following system of equations 
	\begin{equation}
		\begin{aligned}
			\sum_{j=1}^k \eta_j (\phi_{S}(s^{(j)}) - \mu_S ) = 0_{\gH_S}, \quad \sum_{j=1}^k\eta_j = 0
		\end{aligned}
	\end{equation}
	has exactly one solution, i.e., $\eta_j=0$ for all $j\in \{1,\dots, k\}$.
\end{assumption}
Note that the choice of $\kappa_{S}$ can be independent of $\kappa_{XS}$ and $\kappa_{X}$. Since the cardinality of $\Omega_S$ is finite, Assumption \ref{ass:1} is quite mild. A typical choice for $\kappa_S$ is to have linearly independent features $\{\phi_S(s^{(j)})\}_{j=1}^k$. For example, a polynomial kernel with degree $k-1$ would satisfy Assumption \ref{ass:1} for $\Omega_S\subset \mathbb{R}$. The proof is given in Appendix \ref{sec::cok}.  

Then, the following theorem provides insight into the characterization of $\gG_{MP}$.
\begin{theorem}
	\label{theo::gmp}
	Under Assumption \ref{ass:1}, $\gG_{MP}$ is the kernel of the operator $\Sigma_{S(XS)}$, that is,  
	\begin{equation*}
		\mathcal{G}_{MP}=\ker(\Sigma_{S(XS)}).
	\end{equation*}
\end{theorem}

Based on Theorem \ref{theo::gmp},  $\gG_{MP}$ can be found using the relation
\begin{equation*}
	\ker(\Sigma_{S(XS)})=\ran(\Sigma_{(XS)S})^\perp,
\end{equation*}
where $\ran(\Sigma_{(XS)S})$ can be characterized by the generalized eigenvalue problem \citep{hoegaerts2005subset,scholkopf1998nonlinear,  yuan2010reproducing}.

Since $\gH_{S}$ has finite dimension, $\Sigma_{(XS)S}$ is a finite rank operator. Let us say its rank is $m\leq k$. Let $A:\gH_{S}\to \gH_{S}$ be any positive definite linear operator. Then, the first $m$ eigenfunctions of $\Sigma_{(XS)S}A\Sigma_{S(XS)}$, say, $\theta_1, \dots, \theta_m$, span $\ran(\Sigma_{(XS)S})$, that is,
\begin{equation*}
	\ker(\Sigma_{S(XS)})=\spans(\{\theta_1, \dots, \theta_m \})^\perp.
\end{equation*}
Thus, $\gG_{MP}$ can be characterized by a set of eigenfunctions $\{\theta_1, \dots, \theta_m \}$ which allows us to construct an orthogonal projection operator $P$ from $\gH_{XS}$ to $\gG_{MP}$. 

Denote a set of orthonormal bases of  $\ran(\Sigma_{(XS)S})$ by $\{\theta_1',\cdots, \theta_m'\}$. Given a function $g\in \gH_{XS}$, the orthogonal projection operator from $\gH_{XS}$ onto $\ran(\Sigma_{(XS)S})^\perp$ eliminates the components of $g$ in $\ran(\Sigma_{(XS)S})$. Thus, we can construct the following orthogonal projection operator
\begin{equation*}
	P=I-\sum_{j=1}^m \theta_j'\otimes \theta_j',
\end{equation*}
where $I:\gH_{XS}\to \gH_{XS}$ is the identity operator.

For simplicity, the detailed process to estimate $P$ from a given dataset is left to Appendix \ref{sec::prac}.

\paragraph{Remark.} Consider the general CB fairness that seeks to remove the correlation between the sensitive feature map used for prediction and the predicted value. By the observation that $\Cov(\phi_S(S),g(X, S))=\Sigma_{S(XS)}g$, $\ker(\Sigma_{S(XS)})$ is the space whose members are CB fair under the assumption that $\kappa_{XS}$ is composed of $\kappa_S$ and $\kappa_X$, e.g., $\kappa_{XS}=\kappa_X+\kappa_S$. Therefore, the results in the rest of this paper, except the interpretation of tradeoffs, can also be applied to fair regression with CB constraints. The detailed discussion is left to Appendix \ref{app::cb}. In particular, if both the above assumption and Assumption \ref{ass:1} are satisfied in $\gH_{XS}$, MP fairness is equivalent to CB fairness.

\subsection{Optimal fair regression function}
To find the optimal fair regression function, we introduce the optimality condition for Problem \ref{eq::obj}.
\begin{lemma}
	\label{lem::1}
	A function $g^*_{\gG}$ is an optimal solution for Problem \ref{eq::obj} if and only if 
	\begin{equation*}
		\E(Yg(X, S))=\E(g(X, S)g^*_{\gG}(X, S))\quad \forall g\in \gG_{MP}.
	\end{equation*}
\end{lemma}
By introducing the uncentralized covariance operator $\tilde{\Sigma}_{(XS)(XS)}=\E\left[(\phi_{XS}(X, S)\otimes (\phi_{XS}(X, S))\right]$ and a function $h=\E(\phi_{XS}(X, S)Y)$, Lemma \ref{lem::1} tells us that $g^*_{\gG}$ is an optimal solution for Problem \ref{eq::obj} if and only if 
\begin{equation*}
	\langle h,g \rangle_{\gH_{XS}} = \langle \tilde{\Sigma}_{(XS)(XS)}g^*_{\gG}, g\rangle_{\gH_{XS}} \quad \forall g\in \gG_{MP}.
\end{equation*}
Given an orthogonal projection operator $P$ from $\gH_{XS}$ to $\gG_{MP}$, a key insight is that $g^*_{\gG}=Pg^*_{\gH}$ where
$g^*_{\gH}$ can be obtained by solving the following problem
\begin{equation*}
	\langle Ph, g\rangle_{\mathcal{H}_{XS}}=\langle P\tilde{\Sigma}_{(XS)(XS)}P g^*_{\gH}, g\rangle_{\mathcal{H}_{XS}} \quad \forall g\in \gH_{XS},
\end{equation*}
So, we reach the Proposition \ref{pro::opt}.
\begin{proposition}
    \label{pro::opt}
	The optimal MP-fair regression function to Problem \ref{eq::obj} is 
	\begin{equation}
		\label{eq::solution}
		\begin{aligned}
			g^*_{\gG}&=P[P\tilde{\Sigma}_{(XS)(XS)}P]^{\dag}Ph,
		\end{aligned}
	\end{equation} 
	where $(\cdot)^\dagger$ is the Moore-Penrose Inverse of an operator \citep{ groetsch1977generalized,wang2018moore}.
\end{proposition}

Note that if $P$ is an identity operator, the solution \ref{eq::solution}  reduces to $g^*_{\gG}=[\tilde{\Sigma}_{(XS)(XS)}]^\dag h$, which is the least-squares regression function in $\gH_{XS}$.

\subsection{Tradeoff between accuracy and fairness}
\label{sec::trade}
 There are multiple ways of relaxing the MP-fair constraint to control the accuracy-fairness tradeoff. One group of relaxed constraints is imposed on the overall unfairness, e.g., $\text{MPD}(g)\leq \beta$ or $\|\Sigma_{S(XS)}g\|_{\gH_S}\leq \beta$ for some positive real number $\beta$, but such constraints ignore the unfairness for individual group, which weakens their interpretability. Another group of relaxed constraints is imposed on each sensitive group, from which we employ the following relaxed constraint 
 \resizebox{.9\linewidth}{!}{
  \begin{minipage}{\linewidth}
\begin{equation}
	\label{eq::relax_cons}
	\E(g(X, S)|S)-\E(g(X, S))=\alpha \left[\E(g^*(X, S)|S)-\E(g^*(X, S)) \right]
\end{equation}
\end{minipage}
}
where  $g^*$ is the least-squares regression function in $\gH_{XS}$  and $\alpha\in [0,1]$ is a scalar to control the level of unfairness. A larger $\alpha$ results in a higher level of unfairness and $g$ is MP-fair if $\alpha=0$. Thus, the constraint \ref{eq::relax_cons} allows us to scale the unfairness of the least-squares regression function for each group by a scalar $\alpha$ and therefore provides good interpretability. More importantly, we will show that constraint \ref{eq::relax_cons} provides precise control of the accuracy-fairness tradeoff later. 

 To move forward, we present the immediate corollary from Theorem \ref{theo::gmp}.
\begin{corollary}
	\label{co::alpha}
	Given $g_1, g_2\in \gH_{XS}$, under Assumption \ref{ass:1}, $\E(g_1(X, S)|S)-\E(g_1(X, S))= \E(g_2(X, S)|S)-\E(g_2(X, S))$ if and only if  $\Sigma_{S(XS)}g_1=\Sigma_{S(XS)}g_2.$	
\end{corollary}
By Corollary \ref{co::alpha}, it suffices to consider the following relaxed fair regression problem
\begin{equation}
	\begin{aligned}
		\label{eq::trade0}
		\min \quad &\E(Y-g(X, S))^2\\
		\mathrm{s.t. } \quad &\Sigma_{S(XS)}g=\alpha \Sigma_{S(XS)}g^*.
	\end{aligned}
\end{equation}
As $\gH_{XS}=\gG_{MP}\oplus \gG_{MP}^\perp$, a function $g\in \gH_{XS}$ can be written as $g=g_{MP}+g_{MP^\perp}$ where $g_{MP}\in \gG_{MP}$ and $g_{MP^\perp}\in \gG_{MP}^\perp$. Then, the following proposition is the key to solving the Problem \ref{eq::trade0}. 
\begin{proposition}
	\label{pro::1}
	A function $g\in \gH_{XS}$ satisfies $\Sigma_{S(XS)}g=\Sigma_{S(XS)}g^*$ if and only if $g_{MP^\perp} = g^*_{MP^\perp}$.
\end{proposition}
    By Proposition \ref{pro::1}, the optimal solution $g^\alpha$ for Problem \ref{eq::trade0} is of the form $g^\alpha=g_{MP}^\alpha+\alpha	 g^*_{MP^\perp}$, where $g_{MP}^\alpha$ is the optimal solution to the following fair regression problem
\begin{equation}
	\label{eq::trade_1}
	\min_{g\in \gG_{MP}} \E\left(Y-\alpha g^*_{MP^\perp}(X, S)-g(X, S)\right)^2.
\end{equation}
Solving Problem \ref{eq::trade_1} gives the following proposition.
\begin{proposition}
    \label{pro::relax_sol}
	The optimal solution of Problem \ref{eq::trade0} is $g^\alpha=(1-\alpha)g^*_{\gG}+\alpha g^*$
\end{proposition}
Let $L(g)=\E((Y-g(X,S))^2)$. By Proposition \ref{pro::relax_sol}, the following equations allow us to precisely control the tradeoff between fairness and accuracy
	\begin{equation*}
	\begin{aligned}
		L(g^\alpha)&=(1-\alpha)^2L(g^*_{\gG})+(1-(1-\alpha)^2)L(g^*)\\
		\text{MPD}(g^\alpha)&=\alpha \text{MPD}(g^*).
	\end{aligned}
	\end{equation*}	
\paragraph{Remark.} The detailed derivation for this subsection can be found in Appendix \ref{der::trade}. When $\text{MPD}(g^*)>0$, the above equations indicate that $L(g^\alpha)$ is a quadratic function of $\text{MPD}(g^\alpha)$.

\subsection{Performance guarantee}
Besides the explicit expression, the optimal regression function $g^*_{\gG}$ also enjoys a theoretical performance guarantee with respect to MSE.
\begin{proposition}
	\label{pro::2}
	Under Assumption \ref{ass:1}, the MSE of $g^*_{\gG}$ is bounded by
	\begin{equation*}
		\begin{aligned}
			L(g^*_{\gG})\leq& L(g^*) + \langle \tilde{\Sigma}_{(XS)(XS)}g^*_{MP^\perp}, g^*_{MP^\perp}\rangle_{\gH_{XS}},
		\end{aligned}
	\end{equation*}
	where $g^*$ is the optimal regression function in $\gH_{XS}$.
\end{proposition}
In Proposition \ref{pro::2}, the inequality can be obtained by introducing a non-optimal fair regression function $Pg^*$. Note that $g=0_{\gH_{XS}}$ is always a fair regression, so we can claim that $L(g^*_{\gG})\leq \E(Y^2)$. 
Since the term $\langle \tilde{\Sigma}_{(XS)(XS)}g^*_{MP^\perp}, g^*_{MP^\perp}\rangle_{\gH_{XS}}$ measures the  violation of fairness constraints by $g^*$, Proposition \ref{pro::2} shows that the MSE of fair regression function is bounded and the upper bound is related to the unfairness level of $g^*$.

\subsection{Extension}
So far we only consider the fair regression with squared loss function. However, the proposed method can also be applied to other differentiable loss functions in practice. Given a differentiable loss function $\ell$ and the training dataset $\train=\{x_{i}, s_{i}, y_{i}\}_{i=1}^n$,  we consider the following fair regression problem
\begin{equation}
\label{eq::ext}
	\hat{g}^*_{\gG}=\argmin_{g\in \gG_{MP}} \sum_i \ell(y_i, g(x_i,s_i)).
\end{equation}
By the Representer theorem \citep{scholkopf2001generalized}, the above problem is to find $\vw^*_{\gG}$ subject to $\Phi _{XS}\vw^*_{\gG}\in \gG_{MP}$ that minimizes the following objective function
$$J(\vw)=\sum_{i=1}^n \ell\left(y_i, \langle \phi_{XS}(x_i,s_i), \Phi_{XS}\vw\rangle_{\gH_{XS}}\right),$$	
where $\Phi _{XS}$ is the feature matrix of the training data. Given an estimated projection operator $\hat{P}$, we can first find 
$$\vw_\gH = \argmin_{\vw} \sum_{i=1}^n \ell(y_i, \langle \phi_{XS}(x_i,s_i), \hat{P}\Phi_{XS}\vw\rangle_{\gH_{XS}})$$
by optimization techniques, e.g., gradient descent. 

Then, the solution to Problem \ref{eq::ext} is 
$$\hat{g}^*_{\gG} = \hat{P}\Phi_{XS}\vw_\gH.$$

\section{EXPERIMENTS}
    \label{sec::exp}
    	We adapt the experiment settings in \citet{agarwal2019fair} to evaluate the proposed method on  simulated and real-world datasets. The datasets are summarized below:

	\textbf{Synthetic dataset} has  $n$ data points $\{(x_i,s_i,y_i)\}_{i=1}^n$ with $d$-dimension non-sensitive attributes and $e$-dimension sensitive attributes. Specifically, we first generate $\vx_i\sim \mathcal{N}(\vzero_d, \mI_{d\times d})$, $ \vw\sim \mathcal{N}(\vzero_{d+e}, \mI_{(d+e)\times (d+e)})$ and $\epsilon_i \sim \mathcal{N}(0,\rho^2_{noise})$. Then, $\vs_i$  is sampled uniformly at random from $\{0.1, -0.1\}^e$. Next, we set $y_i=[\vx_i, \vs_i]^T\vw+\eps_i$ for linear regression and $y_i=\sin([\vx_i, \vs_i]^T\vw)+\eps_i$ for nonlinear regression (kernel regression case).
	
	\textbf{Adult dataset} \citep{kohavi1996scaling} has 48,842 samples with 14 attributes. We aim to predict the probability that an individual's income exceeds \$50k per year while we keep gender as the sensitive attribute. Our experiments evaluate all methods on a subset of the Adult dataset with 2,000 random samples.
	
	\textbf{Law School dataset} \citep{wightman1998lsac} refers to the Law School Admissions Council’s National Longitudinal Bar Passage Study with 20,649 samples. We aim is to predict a student’s GPA (normalized to [0, 1]) while we keep race as the sensitive attribute. We convert the original race attributes to a single binary attribute, i.e., white or non-white. Our experiments evaluate all methods on a subset of the Law School dataset with 2,000 random samples.
	
	\textbf{Communities \& Crime (C\&C) dataset} \citep{redmond2002data} combines socio-economic, law enforcement, and crime data about communities in the US with  1,994 samples. We aim to predict the number of violent crimes per 100,000	population (normalized to [0, 1]) while we keep race as the sensitive attribute (whether the majority population of the community is white).

	\begin{figure*}[htbp]
		\centering
		\includegraphics[width=\linewidth]{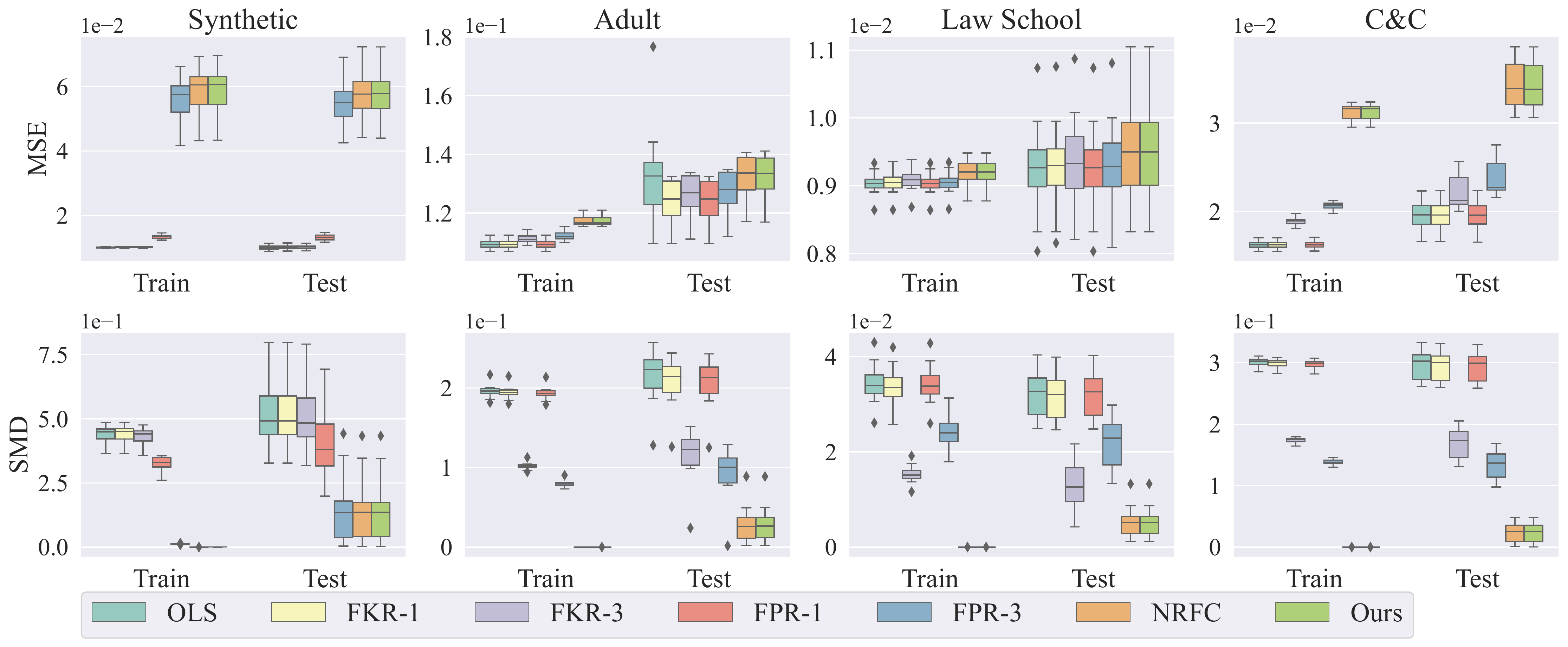}
		\caption{Results of linear regression with single binary sensitive attribute. Figures in the first row show the MSE of different methods, whereas the figures in the second row show the SMD of different methods. The legends FKR-1 and FKR-3 stand for FKR method with regularizer coefficients 10 and 1,000 respectively. Similarly, FPR-1 and FPR-3 stand for FPR method with regularizer coefficients 10 and 1,000 respectively. We also show the experiment results for kernel regression in Appendix \ref{app::supp_exp}.}
		\label{fig::LS_binary_all}
	\end{figure*}
	
	In all experiments, we measure the loss of function $g$ by the empirical MSE
	and the MP disparity by the sum of absolute mean difference (SMD) which is the empirical estimation of $\text{MPD}(g)$ as defined below
	
	\resizebox{.9\linewidth}{!}{
  \begin{minipage}{\linewidth}
  \vspace{-12pt}
	\begin{equation*}
	\text{SMD}(g)=\sum_{j=1}^{k}\left|\frac{\sum_{i=1}^n g(x_i,s_i)\mathbb{I}{(s_i=s^{(j)})}}{\sum_{i=1}^n \mathbb{I}{(s_i=s^{(j)})}}-\frac{\sum_{i=1}^n g(x_i,s_i)}{n}\right|,    
	\end{equation*}
	
	\end{minipage}
}
	where $\mathbb{I}{(\cdot)}$ is the indicator function.
	
	For all datasets, we split the data into two parts, i.e., 80\% for training and 20\% for testing. We discuss the experiments on MP fairness in this section and postpone experiments on CB fairness, DP fairness and regression with other loss functions to Appendix \ref{app::add_exp}. The code is available at \href{https://github.com/shawkui/MP_Fair_Regression}{https://github.com/shawkui/MP\_Fair\_Regression}.

	\subsection{Regression with single binary sensitive attribute}
	\label{sec::exp1}
	
	We first consider regression with single binary sensitive attribute. We claim that MP fairness is equivalent to CB fairness in this setting with proof in Appendix \ref{app::example1}, which allows us to compare the proposed method against the state-of-the-art (SOTA) CB-fair algorithms for regression. Specifically, we compare our method with the ordinary least squares method (OLS), Fair Penalty Regression method (FPR), Fair Kernel Learning method (FKR, \citet{perez2017fair}), and Nonconvex Regression with Fairness Constraints method (NRFC, \citet{komiyama2018nonconvex}) in terms of MSE and SMD, where FKR and NRFC are the SOTA algorithms designed for CB fairness. For regularization-based methods, i.e., FPR and FKR, we evaluate them twice with regularization coefficients $10$ (FPR-1, FKR-1) and $1,000$ (FPR-3, FKR-3) respectively. More details of the baselines and experiment settings can be found in Appendix \ref{exp::baseline}.
	
	The experiment results are summarized in Figure \ref{fig::LS_binary_all}, from which we see that the proposed method can consistently enforce the MP-fair constraint, and its performance is superior to regularization-based methods and competitive with NRFC. Notably, our method can completely remove the algorithmic discrimination on conditional mean for train data. Supplemental Figure \ref{fig::Ker_all} shows our method achieves a smaller MSE than NRFC in kernel regression when both of them reach MP-fairness in train data.

	\begin{figure}[ht]
    \begin{center}
    \centerline{\includegraphics[width=\linewidth]{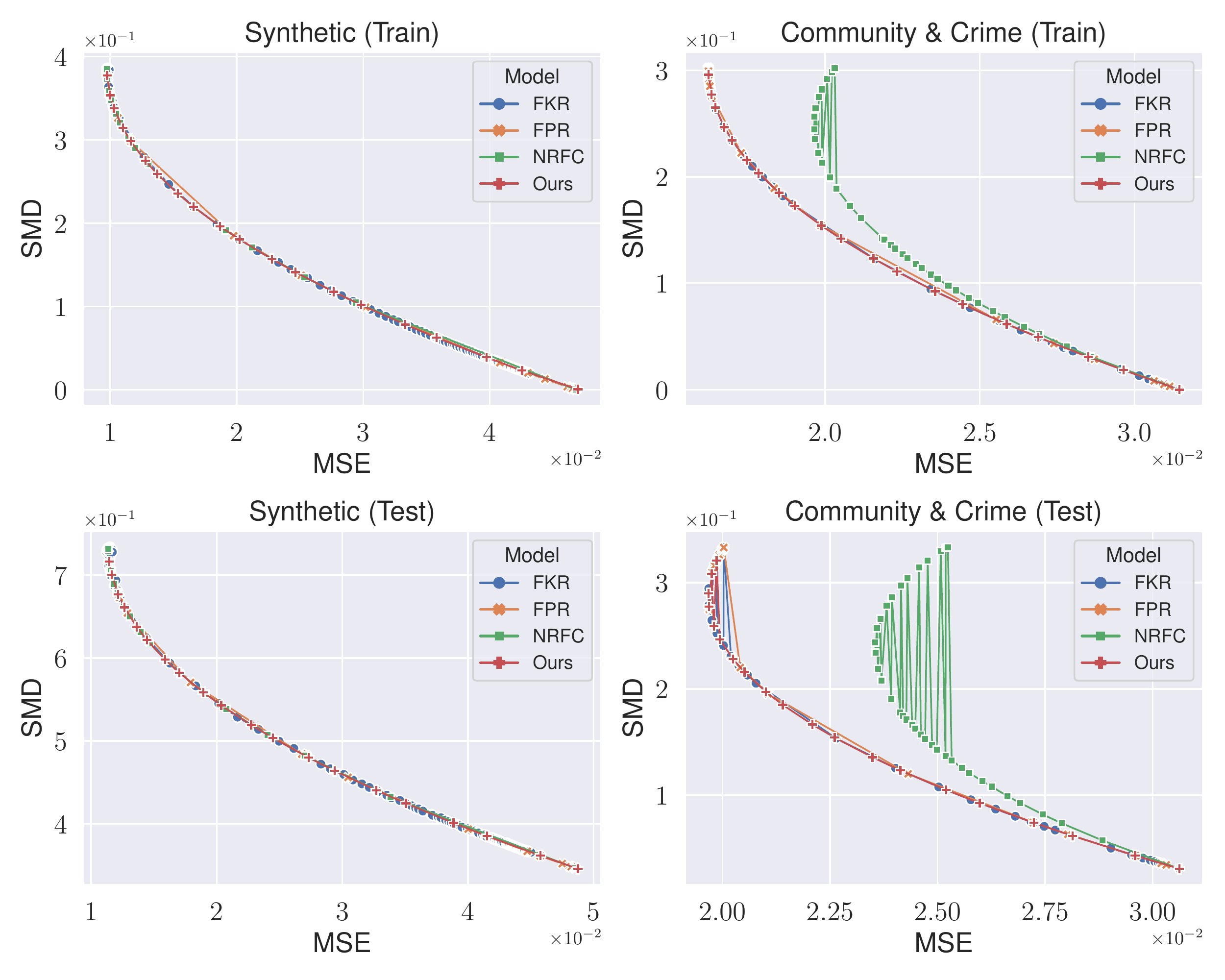}}
    \caption{Results of the fairness-accuracy tradeoff. The first row presents the experiment results for train data whereas the second row shows the experiment results for test data.}
    \label{fig::tradeoff_0}
    \end{center}
    \end{figure}
    
	\begin{figure*}[htbp]
		\centering
		\includegraphics[width=\linewidth]{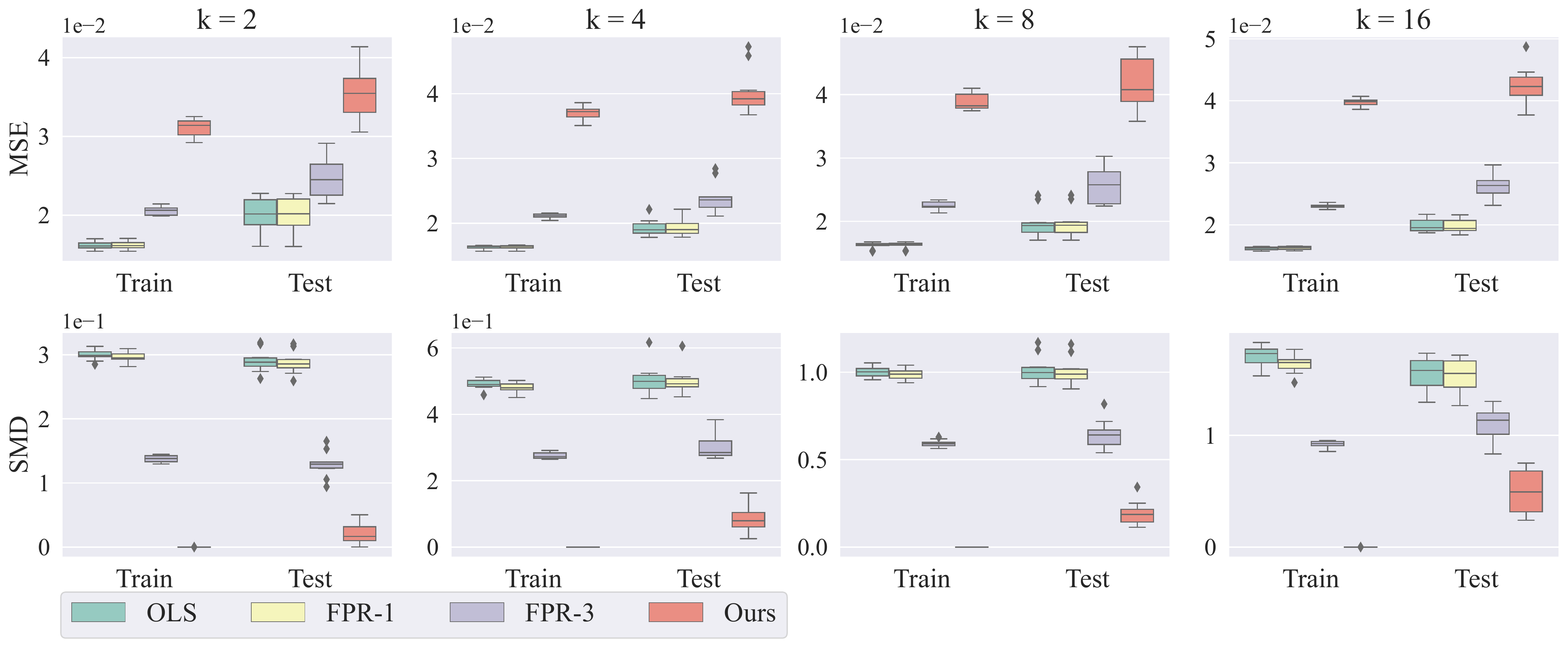}
		\caption{Results of linear regression on Communities \& Crime dataset with multiple binary sensitive attributes. Figures in the first row show the MSE of different methods whereas the figures in the second row show the SMD of different methods. The experiment on kernel regression shows similar results in Appendix \ref{app::supp_exp}.}
		\label{fig::LS_multi_all}
	\end{figure*}
	
	\subsection{Tradeoff between fairness and accuracy}
	\label{sec::exp2}
	We now test the proposed method in Section \ref{sec::trade} on controlling the accuracy-fairness tradeoff, following the setting in Section \ref{sec::exp1}. 
 
	Note that different baselines adopt different metrics and notions for such tradeoff and we only evaluate them in terms of MSE and SMD.	For this purpose, we test the regularization-based methods with fairness regularizer coefficients from $0$ to $10^6$ while for NRFC and the proposed method, we evaluate them with the fairness level parameters from $0$ to~$1$. 
	
    The curves of the fairness-accuracy tradeoff are shown in Figure \ref{fig::tradeoff_0}. As discussed in Section \ref{sec::trade}, the MSE climbs when stricter fairness constraints are imposed. In Figure \ref{fig::tradeoff_0}, the curve of our method coincides with the curves of FKR and FPR, and performs better than the curve of NRFC. When SMD is approaching 0, all methods receive almost the same MSE while NRFC has a higher MSE than other methods on the Communities \& Crime dataset when  weaker fairness constraints are imposed. A similar pattern can be found in supplemental Figure \ref{fig::trade_all} but NRFC and FKR achieve a slightly smaller test MSE sometimes. Although the curves are similar, our method enjoys better explainability and much lower complexity. Unlike other methods which need to solve the regression problem for each level of fairness, our method only solves the regression problem twice and produces a precise tradeoff between fairness and accuracy. 
	
	\subsection{Regression with multiple sensitive attributes}
    \label{sec::exp3}
	As aforementioned, our method can be naturally generalized to regression with multiple sensitive attributes as long as $\kappa_{S}$ satisfies Assumption \ref{ass:1}. In this experiment, we set $\kappa_{S}$ to be a polynomial kernel and choose multiple binary sensitive attributes on the Communities \& Crime dataset. The number of sensitive groups is $k=2^r$ where $r$ is the number of binary sensitive attributes.
	
	In this case, we consider only two baselines: FPR and the OLS since MP-fairness may be not equivalent to CB fairness. Figure \ref{fig::LS_multi_all} depicts the MSE and SMD for different numbers of sensitive attributes, from which we can see that our method can enforce fairness with different numbers of sensitive attributes.

    \begin{figure}[ht]
	\centering
	\includegraphics[width=\linewidth]{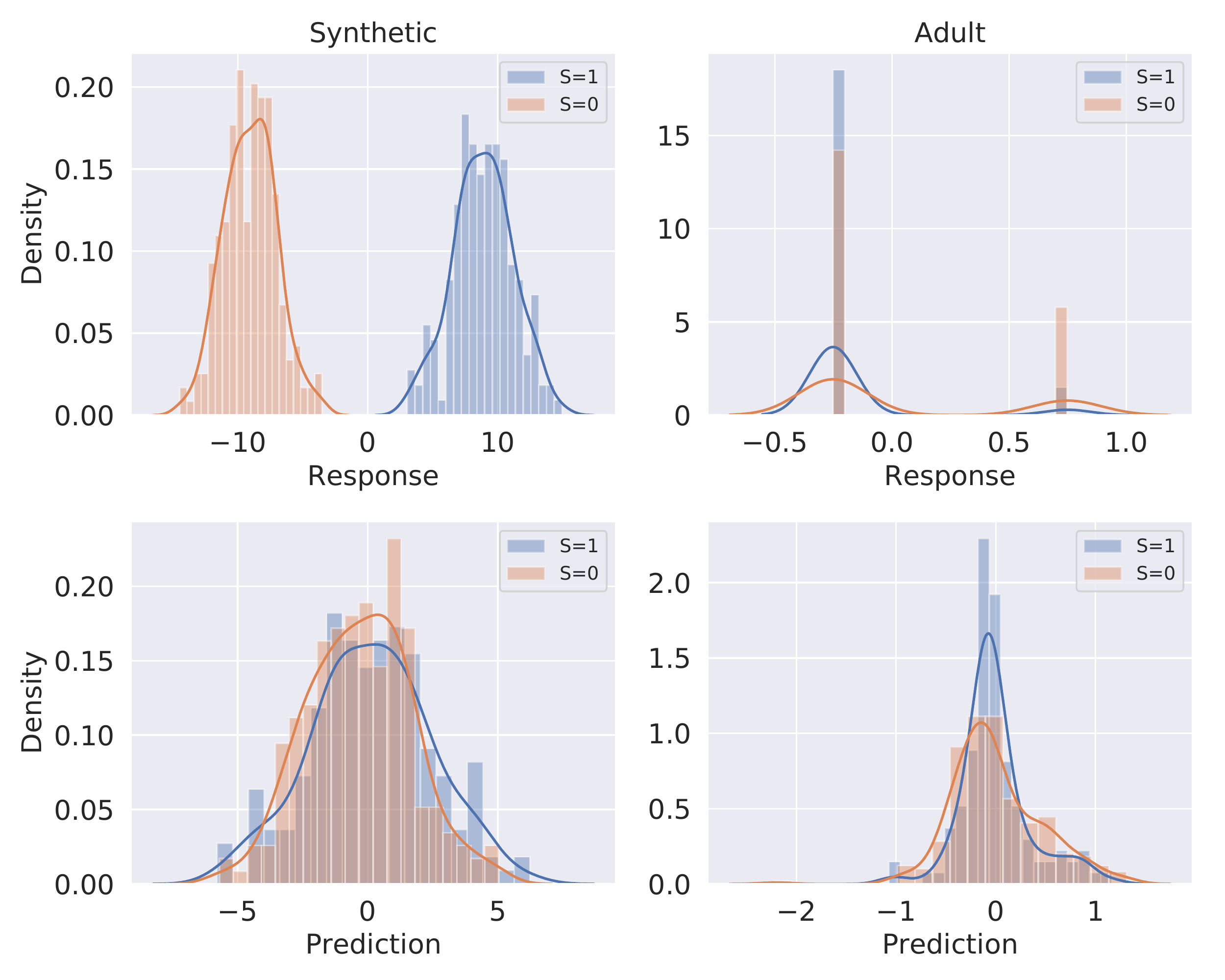}
	\caption{Visualization of centralized response distribution. Both the normalized histograms (bins) and the estimated density (curves) are reported. Figures in the first row show the conditional distribution of response in the test dataset while the figures in the second row show the corresponding conditional distribution of the MP-fair predicted response.}
	\label{fig::dist_0}
    \end{figure}
    
\subsection{Distribution of MP-fair response}
    In this section, we visualize the distribution of response $Y$ and the predicted response $\hat{Y}$ produced by our method on MP-fair regression problem to demonstrate the effect of MP-fairness. Specifically, we consider linear regression with single binary sensitive attribute $S\in\{0,1\}$. Note that to test our method on an extreme case, the synthetic dataset is generated following the linear regression setting with sensitive attribute drawn from $\{-10,10\}$ uniformly so that the distributions of response in two groups are significantly different.

    The results of the Synthetic test dataset and the Adult test dataset are summarized in Figure \ref{fig::dist_0}, from which we can see that the distribution of $\hat{Y}$ conditioning on the sensitive attribute $S$ are similar to each other. This observation agrees with the experiment results in Appendix \ref{app::DP_exp} which says that enforcing MP fairness can significantly reduce the DP disparity.

\section{RELATED WORK}
    \label{sec::rel}
    	\paragraph{Fair regression.} Most prior work on fair regression approximates the optimal fair regression function by data preprocessing or regularizers. Inspired by the two-stage least-squares method used in economics, \citet{komiyama2017two} propose a two-stage algorithm for linear regression that aims to remove the correlation in the dataset, and extend their work to control the level of fairness by employing a nonconvex optimization method \citep{komiyama2018nonconvex}. To provide a general framework for fair regression, \citet{berk2017convex} introduce a family of fairness regularizers for linear regression problems which  enjoy convexity and permit fast optimization.  Similarly, \citet{steinberg2020fast} and \citet{mary2019fairness}  propose to measure the fairness using mutual information and Renyi maximum correlation coefficient respectively  and incorporate the proposed criterion into regularized risk minimization framework. Recently, \citet{scutari2021achieving}  propose a framework for estimating regression models subject to a user-defined level of fairness by introducing a ridge penalty for unfairness. Unlike those works, this paper focuses on the explicit solution to the MP-fair regression problem with both interpretability and theoretical performance guarantees.
	 
    Several works are seeking the explicit solution to the fair regression problem. \citet{calders2013controlling} consider the fair linear regression problem with MP-constraints and provide a closed-form solution using the method of Lagrange multipliers. Based on the connection between least-squares fair regression under Demographic Parity and optimal transport theory, \citet{chzhen2020fairwass} and \citet{gouic2020projection} recently establish the general form of the optimal DP fair regression function and propose a post-processing algorithm that transforms a base estimator of the regression function into a nearly fair one using random smoothing. In the work of \citet{chzhen2022minimax}, the authors consider learning regression function satisfying  $\alpha$-relative DP fair constraint and propose a framework that continuously interpolates between two extreme cases, which is similar to our fairness-accuracy tradeoff method. Other approaches to fair regression include optimization-based methods \citep{oneto2020general}, reduction-based methods \citep{agarwal2018reductions}, and adversary-based methods \citep{chi2021understanding} under some notions of fairness. Unlike them, we focus on MP-fair regression problem in RKHS and derive a closed-form solution by the characterization of fair functional space, which can be extended to covariance-based fairness and other loss functions.

    \paragraph{Kernel methods for algorithmic fairness.} In recent years, kernel methods have drawn increasing attention from the algorithmic fairness community, which can be roughly categorized into two classes. The first class of work aims to employ the kernel method as a regularizer for fairness. \citet{perez2017fair} present the fair kernel ridge regression formulation by incorporating the kernel Hilbert Schmidt independence criterion (KHSIC) as the regularizer on the dependence between the predictor and the sensitive attribute. Similarly, \citet{kim2021learning} propose to learn fair low-rank tensor decompositions by regularizing the Canonical Polyadic Decomposition factorization with the KHSIC. \citet{cho2020fair} develop a kernel density estimation (KDE) methodology for classification problems to quantify the fairness measure as a differentiable function and incorporate it as a regularizer.  Another class of work aims to learn fair representation by leveraging kernel models. In \citet{ grunewalder2021oblivious}, the authors study the relaxed Maximum Mean Discrepancy (MMD) criterion and propose to generate new features that are minimally dependent on the sensitive features while closely approximating the non-sensitive ones. In \citet{okray2019fair}, the authors consider fair regression with binary sensitive attributes and propose to learn fair feature embeddings in kernel space by minimizing the mean discrepancy between the protected group and the unprotected group. In \citet{tan2020learning}, the authors leverage the classical sufficient dimension reduction (SDR) framework to construct fair representations as subspaces of the RKHS under some criterion. Our method differs from those methods from two perspectives: we root in constructing the fair function space and aim to find the explicit solution to the MP fair regression problem.
    
\section{CONCLUSION}
	\label{sec::con}
	In this paper, we have proposed a novel approach for regression under Mean Parity fairness which is appealing both theoretically and practically. By characterizing the space of fair regression functions, we derive a closed-form solution to the fair regression problem which has a simple implementation in practice. The proposed fair function space can also be applied to regression under covariance-based fairness and other loss functions. In addition, our method allows users to control the fairness-accuracy tradeoff systemically and offers a simple interpretation. Experimental results suggest that our approach is promising for applications and improves fairness with multiple sensitive attributes. 
	
	\paragraph{Limitations and future work.} One important direction of future work, and a current challenge is the scalability of the proposed algorithm which is also a common limitation of kernel methods. We remark that many approaches have been proposed to reduce the computational cost of kernel-based algorithms by low-rank matrix approximation \citep{el2014fast, kumar2009sampling} or random projection \citep{cesa2015complexity}, which can also benefit our method. Another valuable direction is to apply our method to other kernel-based models such as Support Vector Machine \citep{noble2006support} and Generalized Linear model \citep{nelder1972generalized}. Other directions of interest include studying the generalization problem of fair algorithms and the characterization of the fair function space for more notions of fairness.

\bibliography{main}
\newpage
\onecolumn
\appendix
\section{PROOFS}
    \label{app::sec1}
    \subsection{Proof of Proposition \ref{pro::mp}}
\label{app::pro_dp}
\begin{proof}
    
    By the fact that $\E(X)=\int_{0}^\infty (1-F(x))dx -\int_{-\infty}^0 F(x)dx$ where $F(X)$ is the  cumulative distribution functions (CDF) of $X$, we have
    \begin{align*}
        \text{MPD}(g)&=\sum_{s\in \Omega_S}|\E(g(X,S)|S=s)-\E(g(X,S))|\\&=\sum_{s\in \Omega_S}\left|\int_\mathbb{R} (F_{g(X,S)|S=s}(t)-F_{g(X,S)}(t))dt\right|\\
        \text{DPD}(g)&=\sum_{s\in \Omega_S} \mathcal{W}_1(g(X,S)|S=s,g(X,S))\\&=\sum_{s\in \Omega_S}\left(\int_\mathbb{R} \left|(F_{g(X,S)|S=s}(t)-F_{g(X,S)}(t))\right|dt\right)
    \end{align*}
    where $F_{g(X,S)|S=s}$ and $F_{g(X,S)}$ are the CDF of $g(X,S)|S=s$ and $g(X,S)$ respectively. 
    
    By the Triangle inequality, we have
            \begin{equation}
                \left|\int_\mathbb{R} (F_{g(X,S)|S=s}(t)-F_{g(X,S)}(t))dt\right| \leq \int_\mathbb{R} \left|(F_{g(X,S)|S=s}(t)-F_{g(X,S)}(t))\right|dt \quad \forall s\in \Omega_S.
            \end{equation}

    So,
    $$\text{MPD}(g)\leq \text{DPD}(g)$$
            
\end{proof}

\subsection{Proof of Theorem \ref{theo::gmp}}
\label{app::gmp}
\begin{proof}
	By the definition of $\ker(\Sigma_{S(XS)})$, a function $g$ is in $\ker(\Sigma_{S(XS)})$ if and only if
	\begin{equation*}
		\Sigma_{S(XS)}g=0_{\gH_S}.
	\end{equation*}
	For a function $g\in \gH_{XS}$, notice that
	\begin{align*}
		\Sigma_{S(XS)}g & =\E_{XS}\left[(\phi_{S}(S)-\mu_s)\otimes (\phi_{XS}(X, S)-\mu_{XS})g\right]           \\
		& =\E_{XS}\left[\langle \phi_{XS}(X, S)-\mu_{XS}, g\rangle_{\mathcal{H}_{XS}}(\phi_{S}(S)-\mu_s) \right] \quad \text{(By the definition of $\otimes$)} \\
		& =\E_{XS}\left[(g(X, S)-\E_{XS}(g(X, S)))(\phi_{S}(S)-\mu_s) \right] \quad \text{(By the reproducing property)}\\
                & =\E_{S}\left[\left(\E_{X}(g(X, S)|S)-\E_{XS}(g(X, S))\right) (\phi_{S}(S)-\mu_s) \right]      \\
		& = \sum_{j=1}^k \sP(S=s^{(j)})\left(\E_{X}(g(X, S)|S=s^{(j)})-\E_{XS}(g(X, S))\right)(\phi_{S}(s^{(j)})-\mu_s)
	\end{align*}
	and 
	$$\sum_{j=1}^k \sP(S=s^{(j)})\left(\E_{X}(g(X, S)|S=s^{(j)})-\E_{XS}(g(X, S))\right)=0$$
	where we use the reproducing property of $\gH_{XS}$ and the definition that $(a\otimes b)c=\langle b,c\rangle_{\mathcal{H}}a$ for $b,c\in \gH$ \citep{gretton2013introduction}.
	
    Note that we assume $\sP(S=s^{(j)})>0$ for all $s^{(j)}\in \Omega_S$ since the sensitive attributes with zero probability don't influence the fairness in practice. Then, under the Assumption \ref{ass:1} that the system of equations 
	\begin{equation}
	\begin{aligned}
	\sum_{j=1}^k \eta_j (\phi_{S}(s^{(j)}) - \mu_S ) = 0_{\gH_S}, \quad 		\sum_{j=1}^k\eta_j = 0
	\end{aligned}
	\end{equation}
	has unique solution $\eta_j=0$ for all $j\in \{1,\dots, k\}$, we can conclude that 
	\begin{align*}
		\Sigma_{S(XS)}g&=0_{\gH_S} \quad \forall  g\in \mathcal{G}_{MP}  \\
		\E_{X}(g(X, S)|S)-\E_{XS}(g(X, S))&=0 \quad \forall g\in \ker(\Sigma_{S(XS)}) .
	\end{align*}
	So, $\mathcal{G}_{MP}=\ker(\Sigma_{S(XS)})$.
\end{proof}

\subsection{Proof of Lemma \ref{lem::1}}
	\label{app::lem1}
	\begin{proof}
		Recall that Problem \ref{eq::obj} considers the following objective
		$$\E(Y-g(X, S))^2,$$
		which is equivalent to
		$$\E(Y^2)-2\E(Yg(X, S))+\E(g(X, S))^2.$$
		
		Denote the optimal regression function of Problem \ref{eq::obj} by $g^*_{\gG}$. Let $\Delta$ be an arbitrary function in $\gG_{MP}$, then $g'=g^*_{\gG}+\Delta$ is a function in $\gG_{MP}$ and
		\begin{align*}
			\E((Y-g'(X, S))^2&=\E(Y^2)-2\E(Yg'(X, S))+\E(g'(X, S))^2\\
			&= \E(Y-g^*_{\gG}(X, S))^2-2\E(Y\Delta(X, S))\\&\quad +2\E(g^*_{\gG}(X, S)\Delta(X, S))+\E(\Delta(X, S))^2.
		\end{align*}
	Note that $g^*_{\gG}$ is an optimal solution if and only if 
    $$\E((Y-g'(X, S))^2\geq \E(Y-g^*_{\gG}(X, S))^2$$
    which is equivalent to
	$$-2\E(Y\Delta(X, S))+2\E(g^*_{\gG}(X, S)\Delta(X, S))+\E(\Delta(X, S))^2\geq 0 \quad \forall \Delta\in \gG_{MP}.$$
	The above inequality holds if and only if
	$$-2\E(Y\Delta(X, S))+2\E(g^*_{\gG}(X, S)\Delta(X, S))=0 \quad \forall \Delta \in \gG_{MP},$$
	which is equivalent to
	$$\E(Y\Delta(X, S))=\E(g^*_{\gG}(X, S)\Delta(X, S)) \quad \forall \Delta \in \gG_{MP},$$
	otherwise, scaling $\Delta$ by a proper scalar yields a contradiction.
	
	\end{proof}

\subsection{Proof of Corollary \ref{co::alpha}}
	\label{app::alpha}
	\begin{proof}
		Given $g_1, g_2\in \gH_{XS}$,  $\E(g_1(X, S)|S)-\E(g_1(X, S))=\E(g_2(X, S)|S)-\E(g_2(X, S))$ indicates that $g_1- g_2 \in \gG_{MP}$, that is, under Assumption \ref{ass:1},
		$$\Sigma_{S(XS)}(g_1-g_2)=0_{\gH_S}.$$
		Rewriting the above equation gives
		$$\Sigma_{S(XS)}g_1= \Sigma_{S(XS)} g_2.$$
	\end{proof}

\subsection{Proof of Proposition \ref{pro::1}}
	\label{app::pro1}
	\begin{proof}
	    A function $g\in \gH_{XS}$ satisfies $\Sigma_{S(XS)}g=\Sigma_{S(XS)}g^*$ if and only if  $\Sigma_{S(XS)}(g-g^*)=0_{\gH_S}.$
		By the definition of $\gG_{MP}$, we have
		\begin{align*}
			\Sigma_{S(XS)}(g-g^*)&=\Sigma_{S(XS)}(g_{MP}-g^*_{MP})+\Sigma_{S(XS)}(g_{MP^\perp}-g^*_{MP^\perp})\\
			&=\Sigma_{S(XS)}(g_{MP^\perp}-g^*_{MP^\perp})\\
			&=0_{\gH_S}.
		\end{align*}
	As $g_{MP^\perp}-g^*_{MP^\perp}\in \gG_{MP}^\perp $, the above equation holds if and only if $g_{MP^\perp}=g^*_{MP^\perp}$.
	\end{proof}

\subsection{Proof of Proposition \ref{pro::2}}
    \label{app::pro2}
	\begin{proof}
		To bound the MSE of $g^*_{\gG}$, we introduce a sub-optimal fair regression function $g'=Pg^*$ where $g^*$ is the optimal regression function in $\gH_{XS}$. Then, the reduction of MSE are
		\begin{align*}
			L(g^*_{\gG})-L(g^*)&\leq L(g')-L(g^*)\\
			&=\E(Y-g'(X,S))^2-\E(Y-g^*(X,S))^2\\
			&=-2\E(Yg'(X, S))+\E(g'(X, S))^2-\left(-2\E(Yg^*(X, S))+\E(g^*(X, S))^2\right) \\
			&=\E(g'(X, S))^2-2\E(Yg'(X, S))+\E(g^*(X, S))^2 \quad \text{(By the optimal condition of $g^*$)}\\
			&=\E(g'(X, S))^2-2\E(g^*(X, S)g'(X, S))+\E(g^*(X, S))^2 \quad \text{(By the optimal condition of $g^*$)}\\
			&=\E(g'(X, S)-g^*(X, S))^2\\
			&=\E(g^*_{MP^\perp}(X,S))^2 \quad \text{(By $g^*=Pg^*+g^*_{MP^\perp}$)}\\
			&=\langle \tilde{\Sigma}_{(XS)(XS)}g^*_{MP^\perp}, g^*_{MP^\perp}\rangle_{\gH_{XS}}.
		\end{align*}
	Therefore, under Assumption \ref{ass:1}, the MSE of $g^*_{\gG}$ is bounded by
	\begin{equation}
		\begin{aligned}
			L(g^*_{\gG})&\leq L(g^*) + \langle \tilde{\Sigma}_{(XS)(XS)}g^*_{MP^\perp}, g^*_{MP^\perp}\rangle_{\gH_{XS}}.
		\end{aligned}
	\end{equation}
	\end{proof}

	\subsection{Proof in Example 1}
	\label{app::example1}
	In this section, we prove that when $S$ is a binary random variable, $\kappa_S(s_i,s_j)=s_is_j$ satisfies Assumption \ref{ass:1}.
	\begin{proof}
	Without loss of generality, we assume that $S\in \{0,1\}$. Since the following system of equations 
 	\begin{equation*}
	\begin{aligned}
	\eta_1(0-\sP(S=1))+\eta_2(1-\sP(S=1))&=0 \\
	\eta_1+\eta_2&=0,
	\end{aligned}
	\end{equation*}
	has a unique solution $\eta_1=\eta_2=0$, Assumption \ref{ass:1} is satisfied.
\end{proof}

\section{IMPLEMENTATION}
	\label{sec::prac}
	In this section, we focus on the estimation of optimal regression function  by solving the empirical approximation of Problem \ref{eq::obj}. Specifically, given the training dataset $\train=\{x_{i}, s_{i}, y_{i}\}_{i=1}^n$, we seek the solution to the following regularized fair regression problem \citep{kadri2010nonlinear, hoegaerts2005subset},
\begin{equation}
	\label{eq::obj_em}
	\min_{g\in \gG_{MP}} \frac{1}{n}\sum_{i=1}^n (y_{i}- g(x_{i}, s_{i}))^2+\frac{\lambda}{n} \left\| g\right\|_{\gH_{XS}},
\end{equation}
where $\lambda\geq 0$ is a real number (regularization coefficient) to control the tradeoff between approximating properties and the smoothness of $g$. Note that when $\lambda =0$, Problem \ref{eq::obj_em} is the estimation of Problem \ref{eq::obj}, but it may be ill-posed depending on $\gH_{XS}$.

To solve Problem \ref{eq::obj_em},  we first show the empirical estimation of $\Sigma_{(XS)S}$ and how to estimate the eigenfunctions of $\Sigma_{(XS)S}A\Sigma_{S(XS)}$, which allows us to construct an orthogonal  projection operator. After that, we derive the closed-form solution for Problem \ref{eq::obj_em} which is the empirical estimation of the optimal fair regression function \ref{eq::solution} when $\lambda = 0$.

\subsection{Empirical estimation of MP-fair function space}
\label{app::B1}
Recall that the feature maps of $\gH_{XS}$ and $\gH_{S}$ are $\phi_{XS}$ and $\phi_{S}$ respectively. Let us define $\bar{\phi}_{XS}(x_i,s_i) = \phi_{XS}(x_i,s_i) - \frac{1}{n}\sum_{j=1}^{n} \phi_{XS}(x_j,s_j)$ and $\bar{\phi}_{S}(s_i)=\phi_{S}(s_i)-\frac{1}{n}\sum_{j=1}^{n}\phi_{S}(s_j)$. Then, the empirical estimation of $\Sigma_{(XS)S}$ is
\begin{equation*}
	\hat{\Sigma}_{(XS)S}=\frac{1}{n}\sum_{i=1}^n \bar{\phi}_{XS}(x_i,s_i)\otimes \bar{\phi}_{S}(s_i).
\end{equation*}
To simplify the derivation, we set $A$ to be the identity operator and focus on $\hat{\Sigma}_{(XS)S}\hat{\Sigma}_{S(XS)} $.

Define the feature matrix $\Phi_{XS}$ and Gram matrix $\mK_{XS}$ as
\begin{align*}
	\Phi_{XS}&=[\phi_{XS}(x_1,s_1), \dots, \phi_{XS}(x_n,s_n)]^T  \\
	\mK_{XS}&=\Phi_{XS}^T\Phi_{XS},
\end{align*}
such that the $i$th column of $\Phi_{XS}$ is $\phi_{XS}(x_i, s_i)$ and the $(i,j)$ entry of $\mK_{XS}$ is $\kappa_{XS}((x_i,s_i), (x_j, s_j))$. Similarly, we denote the feature matrix and Gram matrix of $S$ by $\Phi_S$ and $\mK_{S}$ respectively.

For simplicity, we assume that $\{\bar{\phi}_{XS}(x_i,s_i)\}$ is a set of linearly independent feature maps which ensures that an eigenfunction $\hat{\theta}_l$ is uniquely determined by a set of scalars. In case where  $\{\bar{\phi}_{XS}(x_i,s_i)\}$ are not linearly independent e.g., duplicated data samples, the following process can still be applied since we can get orthonormal bases by removing the duplicated eigenfunctions.

By the observation that $\ran(\hat{\Sigma}_{(XS)S}\hat{\Sigma}_{S(XS)})$ is a subspace of  $\spans(\{\phi_{XS}(x_i,s_i)\}_{i=1}^n)$, the $j$th eigenfunction of $\hat{\Sigma}_{(XS)S}\hat{\Sigma}_{S(XS)}$ can be written as
$$\hat{\theta}_j=\Phi_{XS} \va_j \quad \text{or} \quad \hat{\theta}_j=\bar{\Phi}_{XS}\bar{\va}_j \quad \forall j\in \{1, \dots, m\},$$ 
where $\va_j, \bar{\va}_j\in \sR^n$ are vectors of coefficients and $\bar{\Phi}_{XS}=\Phi_{XS}\mH$ for $\mH=\mI_{n\times n}-\frac{1}{n}\vone_{n\times n}$ \citep{scholkopf1998nonlinear}. 

The generalized eigenvalue $\psi_j$ corresponding to $\hat{\theta}_j$ satisfies
\begin{equation*}
	\label{eq::ev}
	\psi_j \hat{\theta}_j=\hat{\Sigma}_{(XS)S}\hat{\Sigma}_{S(XS)}\hat{\theta}_j.
\end{equation*}
Writing the above equation as a matrix form yields
\begin{equation*}
	\psi_j \bar{\va}_j=\frac{1}{n^2}\bar{\mK}_{S}\bar{\mK}_{XS}\bar{\va}_j,
\end{equation*}
where $\bar{\mK}_S =\mH\mK_{S}\mH $ and $\bar{\mK}_{XS} =\mH\mK_{XS}\mH $.

Thus, $\bar{\va}_j$ is the eigenvector of the matrix $\frac{1}{n^2}\bar{\mK}_{S}\bar{\mK}_{XS}$ and $\va_j=\mH\bar{\va}_j$. Since $\hat{\Sigma}_{(XS)S}\hat{\Sigma}_{S(XS)}$ is self-adjoint, the first $m$ eigenfunctions are orthogonal. So, we can normalize the eigenfunctions to construct a set of orthonormal bases of $\ran(\hat{\Sigma}_{(XS)S})$. A more detailed derivation can be found in Appendix \ref{app::eign}.

\subsection{Construction of projection operator}
With some abuse of notation, we denote a set of orthonormal bases of  $\ran(\Sigma_{(XS)S})$ by $\{\theta_1,\cdots, \theta_m\}$ and its estimation by $\{\hat{\theta}_1,\cdots, \hat{\theta}_m\}$ where $\hat{\theta}_j=\Phi_{XS}\va_j$  to avoid complicated symbols. Given a function $g\in \gH_{XS}$, the orthogonal projection operator from $\gH_{XS}$ onto $\ran(\Sigma_{(XS)S})^\perp$ eliminates the components of $g$ in $\ran(\Sigma_{(XS)S})$. Thus, we can construct the following orthogonal projection operator
\begin{equation*}
	P=I-\sum_{j=1}^m \theta_j\otimes \theta_j,
\end{equation*}
where $I:\gH_{XS}\to \gH_{XS}$ is the identity operator.

So, the estimation of $P$ can be written as 
\begin{equation}
    \hat{P}=I-\sum_{j=1}^m \hat{\theta}_j\otimes \hat{\theta}_j
\end{equation}
Note that given $g=\Phi_{XS}\vc$, the projection of $g$ on $\gG_{MP}$ is 
\begin{equation*}
	\begin{aligned}
		\hat{P}g&=g-\sum_{j=1}^{m}\langle g, \hat{\theta}_j\rangle_{\mathcal{H}_{XS}}\hat{\theta}_j=\Phi_{XS}\mP \vc,
	\end{aligned}
\end{equation*}
where $\mP=(\mI_{n\times n}-\sum_{j=1}^{m}\va_m\va_m^T\mK_{XS})$. 

\subsection{Estimation of fair regression function}
\label{sec::s43}
Given an orthogonal projection operator estimation $\hat{P}$, the optimal solution to Problem \ref{eq::obj_em} is $\hat{g}^*_{\gG}=\hat{P}\hat{g}^*_\gH$ where $\hat{g}^*_\gH$ can be obtained by solving the following problem
\begin{equation*}
	\min_{g\in \gH_{XS}} \frac{1}{n}\left(y_i-\langle \phi_{XS}(x_i,s_i), \hat{P}g\rangle\right)^2+\frac{\lambda}{n}\left\| \hat{P}g\right\|_{\gH_{XS}}.
\end{equation*}

By the Representer theorem \citep{scholkopf2001generalized}, $\hat{g}_\gH$ is of the form $\hat{g}_\gH=\Phi_{XS}\vw_\gH$ for $\vw_\gH\in \sR^{n}$. So, it suffices to minimize the following objective function
\begin{equation*}
	\begin{aligned}
		J(\vw)=&\vw^T\mP^T\mK_{XS}\mK_{XS}\mP\vw-2\mY^T\mK_{XS}\mP\vw\\
		& + \mY^T\mY+\lambda \vw^T\mP^T\mK_{XS}\mP\vw,
	\end{aligned}
\end{equation*}
where $\mY=[y_1, \dots, y_n]$ is a vector in $\sR^{n}$.

Since $J(\vw)$ is convex, it has a minimizer. Setting $\frac{\partial J}{\partial \vw}$ to zero yields
\begin{equation*}
	\vw_\gH=(\mP^T\mK_{XS}\mK_{XS}\mP+\lambda \mP^T\mK_{XS}\mP)^{\dagger}\mP^T\mK_{XS}\mY.
\end{equation*}
So, the optimal fair regression function is $\hat{g}^*_{\gG}=\Phi_{XS}\vw^*_{\gG}$ where
$$\vw^*_{\gG}=\mP (\mP^T\mK_{XS}\mK_{XS}\mP+\lambda \mP^T\mK_{XS}\mP)^{\dagger}\mP^T\mK_{XS}\mY.$$

\textbf{Example: fair linear regression}.  Consider the fair linear regression problem with single binary sensitive attribute. The kernels are 
\begin{align*}
	\kappa_{S}(s_i,s_j)& =s_is_j\\
	\kappa_{X}(x_i,x_j)& =x_i^Tx_j\\
	\kappa_{XS}((x_i,s_i),(x_j,s_j)) & =\kappa_{S}(s_i,s_j)+\kappa_{X}(x_i,x_j)
	.	\end{align*}
We prove that the above setting satisfies Assumption \ref{ass:1} in Appendix \ref{app::example1}, which implies that MP fairness is equivalent to CB fairness in this example. Let $\lambda=0$.
The optimal fair regression function is $$\hat{g}^*_{\gG}=\Phi_{XS}\mP (\mK_{XS}\mP)^{\dagger}\mY$$
and the fitted value of $\mY$ is
$$\hat{\mY}=\mK_{XS}\mP(\mK_{XS}\mP)^\dag \mY.$$

\section{DERIVATIONS AND DISCUSSIONS}
	\label{sec::der}

\subsection{Relation between MP fairness and CB fairness}
\label{app::cb}
In this section, we discuss general CB fairness and its relation to MP fairness. We first provide the following assumption
\begin{assumption}
    \label{ass::2}
    Assume the $\kappa_{XS}$ is composed of $\kappa_S$ and $\kappa_X$.
\end{assumption}
which is the assumption in ordinary kernelized regression problem where $S$ and $X$ are mapped to $\phi_S(S)$ and $\phi_{X}(X)$ respectively. As discussed in the work of \citet{komiyama2018nonconvex} and \citet{perez2017fair}, the general CB fairness seeks to remove the correlation between $S$ and $g(X, S)$ on the (possibly infinite) representation space. Specifically, the CB fairness requires that the regression function $g\in \gH_{XS}$ achieves $\Cov(\phi_S(S),g(X, S))=0_{\gH_S}$ under Assumption \ref{ass::2}. By the definition of  $\Cov(\phi_S(S),g(X, S))$, we have
\begin{equation}
\label{eq::cov}
\begin{aligned}
    \Cov(\phi_S(S),g(X, S))&=\E\left[(g(X, S)-\E(g(X, S)))(\phi_S(S)-\mu_S)\right]\\
    &=\E\left[\langle\phi_{XS}(X, S)-\mu_{XS}, g \rangle_{\gH_{XS}}(\phi_S(S)-\mu_S)\right]\\
    &=\E\left[(\phi_S(S)-\mu_S)\otimes (\phi_{XS}(X, S)-\mu_{XS}) \rangle\right]g\\
    &=\Sigma_{S(XS)}g,
\end{aligned}
\end{equation} 
where we use the reproducing property of $\gH_{XS}$ and the definition that $(a\otimes b)c=\langle b,c\rangle_{\mathcal{H}}a$ for $b,c\in \gH$ \citep{gretton2013introduction} to derive this result.

Equation \ref{eq::cov} claims that a function $g\in \gH_{XS}$ is CB-fair if and only if $g$ is in $\ker(\Sigma_{S(XS)})$. Since the proposed method solves the fair regression problem by the characterization of $\ker(\Sigma_{S(XS)})$, it can also be applied to CB fairness under Assumption \ref{ass::2}. In particular, if both Assumption \ref{ass:1} and Assumption \ref{ass::2} are satisfied, MP fairness is equivalent to CB fairness.

\subsection{Derivation of equations in Section \ref{sec::trade}}
\label{der::trade}
Solving Problem \ref{eq::trade_1} gives
\begin{equation*}
	\begin{aligned}
		g^\alpha_{MP}&=P[P\tilde{\Sigma}_{(XS)(XS)}P]^{\dag}P(h-\alpha \tilde{\Sigma}_{(XS,XS)}g^*_{MP^\perp})\\
		& =g^*_{\gG}-\alpha P[P\tilde{\Sigma}_{(XS)(XS)}P]^{\dag}P \tilde{\Sigma}_{(XS,XS)}g^*_{MP^\perp}.
	\end{aligned}
\end{equation*}
As $g^*_{MP^\perp}=(I-P)g^*$ where $I$ is the identity operator,  we have
\begin{align*}
	P[P\tilde{\Sigma}_{(XS)(XS)}P]^{\dag}&P \tilde{\Sigma}_{(XS,XS)}g^*_{MP^\perp}\\&=P[P\tilde{\Sigma}_{(XS)(XS)}P]^{\dag}P \tilde{\Sigma}_{(XS,XS)}g^*-P[P\tilde{\Sigma}_{(XS)(XS)}P]^{\dag}P \tilde{\Sigma}_{(XS,XS)}Pg^*
\end{align*}
where the first term equals to $g^*_{\gG}$ by the property that $\langle g, \tilde{\Sigma}_{(XS,XS)}g^*\rangle_{\gH_{XS}}= \langle g,h\rangle_{\gH_{XS}}$ for all $g\in \gH_{XS}$, and the second term equals to $Pg^*$ since it's the optimal solution of $\min_{g\in \gG_{MP}} \E(Pg^*-g)^2$.

Therefore, we get
$$g^\alpha_{MP}=(1-\alpha)g^*_{\gG}+\alpha Pg^*.$$	
Alternatively, we can show the above equation using the fact that $g^\alpha = g^*$ when $\alpha=1$. 

Thus, the optimal solution to Problem \ref{eq::trade0} is $g^\alpha=(1-\alpha)g^*_{\gG}+\alpha g^*$. 

Now, we turn to the MSE of $g^\alpha$. We have
\begin{equation}
    \begin{aligned}
    L(g^\alpha)&=\E(Y-g^\alpha(X,S))^2\\
    &=\E(Y-(1-\alpha)g^*_{\gG}(X,S)-\alpha g^*(X,S))^2\\
    &=\E((1-\alpha)(Y-g^*_{\gG}(X,S))+\alpha (Y-g^*(X,S)))^2\\
    &=\E((1-\alpha)(Y-g^*_{\gG}(X,S)))^2+\E(\alpha (Y-g^*(X,S)))^2\\&\quad +2\E((1-\alpha)(Y-g^*_{\gG}(X,S))(\alpha (Y-g^*(X,S))))\\
    &=\E((1-\alpha)(Y-g^*_{\gG}(X,S)))^2+\E(\alpha (Y-g^*(X,S)))^2\\&\quad +2\alpha(1-\alpha)\E((Y-g^*_{\gG}(X,S))(Y-g^*(X,S)))\\
    &=(1-\alpha)^2\E(Y-g^*_{\gG}(X,S)))^2+(1-(1-\alpha)^2)\E((Y-g^*(X,S)))^2\\
    &=(1-\alpha)^2L(g^*_{\gG})+(1-(1-\alpha)^2)L(g^*)\\
    &=\alpha^2 (L(g^*_{\gG})-L(g^*)) - 2\alpha (L(g^*_{\gG})-L(g^*)) + L(g^*_{\gG}),\\
    &=(1-\alpha)^2L(g^*_{\gG})+(1-(1-\alpha)^2)L(g^*)\\
		\text{MPD}(g^\alpha)&=\alpha \text{MPD}(g^*).
    \end{aligned}
\end{equation}
    since $\E(Y(Y-g^*(X,S)))=\E(Y-g^*(X,S))^2$ and $\E(g^*_{\gG}(X,S)(Y-g^*(X,S)))=0$.

\subsection{Estimating eigenfunctions and orthonormal bases}
\label{app::eign}
Now we provide detailed derivation about finding the eigenfunctions of $\hat{\Sigma}_{(XS)S}A\hat{\Sigma}_{S(XS)}$. 

Recall that
\begin{equation*}
	\hat{\Sigma}_{(XS)S}=\frac{1}{n}\sum_{i=1}^n \bar{\phi}_{XS}(x_i,s_i)\otimes \bar{\phi}_{S}(s_i).
\end{equation*}
Let $A$ be the identity operator, and we get
\begin{align*}
	\hat{\Sigma}_{(XS)S}A\hat{\Sigma}_{S(XS)} & =\left(\frac{1}{n}\sum_{i=1}^n \bar{\phi}_{XS}(x_i,s_i)\otimes \bar{\phi}_{S}(s_i) \right) \left(\frac{1}{n}\sum_{i=1}^n \bar{\phi}_{S}(s_i)\otimes \bar{\phi}_{XS}(x_i,s_i)\right)       \\
	& =\frac{1}{n^2}\sum_{i=1}^{n}\sum_{j=1}^{n} {\bar{\phi}_{XS}(x_i,s_i)\otimes \bar{\phi}_{S}(s_i)\bar{\phi}_{S}(s_j)}\otimes \bar{\phi}_{XS}(x_j,s_j)                                  \\
	& =\frac{1}{n^2}\sum_{i=1}^{n}\sum_{j=1}^{n}    {\langle  \bar{\phi}_{S}(s_i), \bar{\phi}_{S}(s_j) \rangle_{\mathcal{H}_{S}} \bar{\phi}_{XS}(x_i,s_i)}\otimes \bar{\phi}_{XS}(x_j,s_j) \\
	& =\frac{1}{n^2}\sum_{i=1}^{n}\sum_{j=1}^{n} \bar{\kappa}_{S}(s_i,s_j) \bar{\phi}_{XS}(x_i,s_i)\otimes \bar{\phi}_{XS}(x_j,s_j).
\end{align*}

For simplicity, we assume that $\{\bar{\phi}_{XS}(x_i,s_i)\}$ is a set of independent feature maps which ensures that $\hat{\theta}_l$ is uniquely determined by a set of scalars. In case where  $\{\bar{\phi}_{XS}(x_i,s_i)\}$ are not independent e.g., duplicated data samples, the following process can still be applied since we can get orthonormal bases by removing the duplicated eigen functions.

Since $\hat{\Sigma}_{(XS)S}\hat{\Sigma}_{S(XS)}$ is in  $\ran(\{\bar{\phi}_{XS}(x_i,s_i)\}_{i=1}^n)$, the $l^{th}$ eigenfunction  of $\hat{\Sigma}_{(XS)S}\hat{\Sigma}_{S(XS)}$ can be written as
\begin{equation*}
	\hat{\theta}_l=\sum_{i=1}^{n} \bar{a}_{l_i}\bar{\phi}_{XS}(x_i,s_i).
\end{equation*}

By the definition of eigenfunction, we get
\begin{equation}
	\label{eq::eign}
	\psi_l \hat{\theta}_l=\hat{\Sigma}_{(XS)S}\hat{\Sigma}_{S(XS)}\hat{\theta}_l.
\end{equation}

Observe that
\begin{equation*}
	\langle \bar{\phi}_{XS}(x_i,s_i), \sum_{j=1}^{n} \bar{a}_{l_j}\bar{\phi}_{XS}(x_j,s_j) \rangle_{\mathcal{H}_{XS}}=\sum_{j=1}^{n} \bar{a}_{l_j}\bar{\kappa}_{XS}\left((x_i,s_i), (x_j,s_j)\right),
\end{equation*}
where $\bar{\kappa}_{XS}\left((x_i,s_i), (x_j,s_j)\right)$ is the $(i,j)$ entry of the matrix $\bar{\mK}_{XS}=\mH\mK_{XS}\mH$ with Gram matrix $\mK_{XS}$ and $\mH=\mI_{n\times n}-n^{-1}\vone_{n\times n}$. Thus, we get
\begin{equation*}
	\hat{\Sigma}_{(XS)S}\hat{\Sigma}_{S(XS)}\hat{\theta}_l=\frac{1}{n^2}\sum_{i=1}^{n}\sum_{j=1}^{n} \beta_{l_j} \bar{\kappa}_{S}(s_i,s_j) \bar{\phi}_{XS}(x_i,s_i), 
\end{equation*}
where $\beta_{l_j}=\sum_{r=1}^{n} \bar{a}_{l_r}\bar{\kappa}_{XS}\left((x_j,s_j), (x_r,s_r)\right)$.

By Equation \ref{eq::eign}, it suffices to solve
\begin{equation*}
	\lambda_{l}\bar{a}_{l_i}=\frac{1}{n^2} \sum_{j=1}^{n}  \bar{\kappa}_{S}(s_i,s_j) \sum_{r=1}^{n} \bar{a}_{l_r}\bar{\kappa}_{XS}\left((x_j,s_j) (x_r,s_r)\right).
\end{equation*}
Writing the above equation as a matrix equation yields
\begin{equation*}
	\psi_l \bar{\va}_l=\frac{1}{n^2}\bar{\mK}_{S}\bar{\mK}_{XS}\bar{\va}_l,
\end{equation*}
where $\va_l = [\bar{a}_{l_1}, \dots , \bar{a}_{l_n}]$ is a column vector in $\sR^{n}$.

Thus, $\va_l$ is the eigenvector of the matrix $\frac{1}{n^2}\bar{K}_{S}\bar{K}_{XS}$.
Let
\begin{equation*}
	\va_l=\mH\bar{\va}_l.
\end{equation*}
The $l^{th}$ eigenvector can be rewritten as
\begin{equation*}
	\hat{\theta}_l=\sum_{i=1}^{n} a_{l_i}\phi_{XS}(x_i,s_i).
\end{equation*}

Since $\hat{\Sigma}_{(XS)S}\hat{\Sigma}_{S(XS)}$ is self-adjoint, the first $m$ eigenfunctions are orthogonal. So, we can normalize the eigenfunctions to construct a set of orthonormal bases of $\ran(\hat{\Sigma}_{(XS)S})$.

\subsection{Choice of kernel}
\label{sec::cok}
For Mean Parity Fair Regression, the choice of $\kappa_S$ is independent of $\kappa_{XS}$ and $\kappa_X$ as long as $\phi_S$ satisfies Assumption~1. Here we show that a polynomial kernel with degree $k-1$ would satisfy Assumption 1 for $\Omega_S\subseteq \mathbb{R}$, i.e., $S$ is a scalar variable.

Consider a polynomial kernel with degree of $k-1$, i.e., $\kappa_S(s_1,s_2)=(1+s_1s_2)^{k-1}$. The feature map $\phi_S(s)$ is
$$\phi_S(s)=[c_0, c_1s, c_2s^2, \cdots, c_{k-1}s^{k-1}]$$
where $c_i=\sqrt{{k-1 \choose i}}$ according to the binomial theorem.

Now we show that $\{\phi_S(s^{(j)})\}_{j=1}^k$ is a set of linearly independent feature maps by showing the following problem has no non-zero solution
\begin{equation}
\label{eq::linear}
  \sum_{j=1}^kw_j\phi_S(s^{(j)})=0_{\gH_S}  
\end{equation}

Equation~\ref{eq::linear} is equivalent to 
$$ \begin{bmatrix}
c_0 & c_0 & c_0 & \cdots & c_0 \\
c_1(s^{(1)})^1 & c_1(s^{(2)})^1 & c_1(s^{(3)})^1 & \cdots & c_1(s^{(k)})^1 \\
c_2(s^{(1)})^2 & c_2(s^{(2)})^2 & c_2(s^{(3)})^2 & \cdots & c_2(s^{(k)})^2 \\
\vdots & \vdots & \vdots & \vdots & \vdots \\
c_{k-1}(s^{(1)})^{k-1} & c_{k-1}(s^{(2)})^{k-1} & c_{k-1}(s^{(3)})^{k-1} & \cdots & c_{k-1}(s^{(k)})^{k-1} \\
\end{bmatrix} 
\begin{bmatrix}
w_1\\w_2\\w_3\\\vdots\\w_k 
\end{bmatrix} = \begin{bmatrix}
0\\0\\0\\\vdots\\0
\end{bmatrix}
$$
We can simplify  the above problem to

$$  \underbrace{\begin{bmatrix}
1 & 1 & 1 & \cdots & 1 \\
(s^{(1)})^1 & (s^{(2)})^1 & (s^{(3)})^1 & \cdots & (s^{(k)})^1 \\
(s^{(1)})^2 & (s^{(2)})^2 & (s^{(3)})^2 & \cdots & (s^{(k)})^2 \\
\vdots & \vdots & \vdots & \vdots & \vdots \\
(s^{(1)})^{k-1} & (s^{(2)})^{k-1} & (s^{(3)})^{k-1} & \cdots & (s^{(k)})^{k-1} \\
\end{bmatrix}}_{\mV} 
\begin{bmatrix}
w_1\\w_2\\w_3\\\vdots\\w_k 
\end{bmatrix} = \begin{bmatrix}
0\\0\\0\\\vdots\\0
\end{bmatrix}
$$

Since the matrix $\mV$ is a Vandermonde Matrix, it has determinant $\det(\mV)=\prod\limits_{1\leq i<j\leq k}(s^{(j)}-s^{(i)})\neq 0$. Therefore, the above problem has no non-zero solution and  $\{\phi_S(s^{(j)})\}_{j=1}^k$ is a set of linearly independent features. Thus, a polynomial kernel with degree $k-1$ would satisfy Assumption 1 for scalar-valued  $S$.

For sensitive attributes with non-scalar value, a modified polynomial kernel that first maps $S$ to scalar value and then computes the features by the standard polynomial kernel can be well adopted.

\section{EXPERIMENTS DETAILS}
    \label{app::exp}
    
\subsection{Baselines}
\label{exp::baseline}
The details of the baselines used in the experiments are summarized below: 
\begin{itemize}
    \item Constant Prediction: a regression function with a constant outcome that minimizes the MSE. It achieves MP, DP and CB fairness.
    \item Ordinary Least Squares: the standard linear regression model without regularizers.
    \item Kernel Ridge Regression: the standard kernel regression \citep{welling2013kernel} method with regularizers.
    \item Fair Penalty Regression: a regression model with MP-fair regularizers. Derivation can be found in Appendix \ref{penalty}.
    \item Fair Kernel Learning  \citep{perez2017fair}: a regularizer-based method aims to eliminate the covariance between the predicted value and the sensitive attributes. The implementation is borrowed from \href{https://isp.uv.es/soft_regression.html}{https://isp.uv.es/soft\_regression.html}.
    \item Nonconvex Regression with Fairness Constraints \citep{komiyama2018nonconvex}: a nonconvex optimization method aims to control the correlation between the predicted value and the sensitive attributes. Note that the optimization process is applied only when the target CB disparity is set to be larger than 0, otherwise, NRFC is reduced to a data preprocessing method. We adapt the official implementation from \href{https://github.com/jkomiyama/fairregresion}{https://github.com/jkomiyama/fairregresion}.
    \item Reduction Based Algorithm \citep{agarwal2019fair}: a reduction based method aims to achieve DP fairness for a randomized predictor using discretization. We adapt the official implementation from \href{https://github.com/steven7woo/fair_regression_reduction}{https://github.com/steven7woo/fair\_regression\_reduction}.
    
\end{itemize}

\subsection{Experiment settings}
\label{exp::set}

The detailed settings in each experiment are summarized below:
\begin{itemize}
    \item \textbf{Data preprocessing. } For all experiments, both response values in train data and test data are centralized using the mean of training response values.
    
    \item \textbf{Linear regression. } For synthetic dataset, we choose $n=2,000$, $d=5$ and $e=1$ for regression with single sensitive attribute. The variance of noise $\rho^2_{noise}$ is set to be $0.1$. For the proposed method and FPR, we set the kernel of sensitive attributes as the polynomial kernel. All methods focus on the unregularized least-squares problem, i.e, $\lambda =0$. We test FKR with fairness regularizer coefficients $10$ and $1,000$ which are represented by FKR-1 and FKR-3 respectively. Similarly, We test FPR with coefficients of fairness regularizer $10$ and $1,000$ which are represented by FPR-1 and FPR-3 respectively.  Note that for NRFC, it defaults to fit linear regression with intercept. So, when evaluating other methods, we add a column of ones to $X$ to match the setting of NRFC. Other settings for the hyper-parameters in the baselines follow the default settings of their corresponding papers. We run each method 10 times.

    \item \textbf{Kernel regression. } For the proposed method, we set $\kappa_S$ to be polynomial kernel while all other kernels are set to be Radial Basis Function (RBF) Kernel with $\gamma = 0.1$. We focus on the regularized least-squares problem with $\lambda =1$. Other settings for the hyper-parameters in the baselines are the same as the settings in the linear regression experiment.
    
    \item \textbf{Tradeoff.} The proposed method is evaluated with $\alpha =[0, 1/50, 2/50, \dots, 1]$. For FKR and FPR, we alter the coefficient of fairness regularizer from $0$ to $10^6$. Moreover, we run NRFC with $\zeta$, the parameter for the level of fairness from 0 to 1. Note that except $\alpha$, all other parameters need to be tuned carefully since the relation between fairness and accuracy is hard to interpret (sometimes a small change in the fairness parameter will make a dramatic change to the loss while sometimes the change is negligible). In particularly, the values of $\zeta$ concentrate in $[0,0.1]$ and even $[0,0.01]$. For the regularizer coefficient of FKR and FPR, the values concentrate in $[10^2, 10^4]$. To make the figures clear, we plot a subset of experiment results in Figure \ref{fig::tradeoff_0} and Figure \ref{fig::trade_all} by subsampling $1/5$ of the results uniformly.
    
    \item \textbf{Multiple sensitive attributes.} For regression with multiple sensitive attributes on the Communities and Crime dataset, we choose race, medIncome, householdsize and medFamInc as the sensitive attributes sequentially. For medIncome, householdsize and medFamInc, we convert them to binary attributes by whether their values are larger than 0.5.

\end{itemize}

\subsection{Fair penalty regression}
    \label{penalty}
	In this section, we derive an FPR model for MP fairness using the framework of \cite{perez2017fair} which is used as a baseline in our experiment. For the FPR model, the key point is to find function $Q(g(X, S),S)$ which measures the level of MP-fairness of a regression function. Notice that a function $g\in \gH_{XS}$ satisfies MP fairness if and only if its projection onto $\gG_{MP}$ is itself, i.e., $g-Pg=0_{\gH_{XS}}$. So, we set
	$$Q(g(X, S),S)=\left\| g-Pg\right \|_{\mathcal{H}_{XS}}.$$
 Given the training dataset $\train=\{x_{i}, s_{i}, y_{i}\}_{i=1}^n$, we seek the solution of the following regularized optimization problem,
	\begin{equation}
	\label{eq::obj_eta}
	\min_{g\in \gG_{MP}} \frac{1}{n}\sum_{i=1}^n (y_{i}- g(x_{i}, s_{i}))^2+\frac{\lambda}{n} \left\| g\right\|_{\gH_{XS}} +\frac{\zeta}{n}\left\| g-Pg\right \|_{\mathcal{H}_{XS}}. 
	\end{equation}
 where $\zeta\geq 0$ is the parameter to control the level of fairness. 
 
	By the Representer theorem, the optimal solution $g^*$ is of the form $\Phi_{XS}\vw$. So, we need to solve the problem
	\begin{equation}
	\min_{g\in \gG_{MP}} \mY^T\mY-2\mY^T\mK_{XS}\vw+\vw^T\mK_{XS}\mK_{XS}\vw+\lambda \vw^T\mK_{XS}\vw + \zeta \vw^T \mK_{XS}\mA^T \mK_{XS} \mA\mK_{XS}\vw,
	\end{equation}
	where $\mA=\sum_{j=1}^{m}\va_j\va_j^T$. Since the above problem is convex, its has a solution 
	\begin{align*}
		\vw&=\left(\mK_{XS}\mK_{XS}+\lambda \mK_{XS} +\zeta\mK_{XS}\mA^T \mK_{XS} \mA\mK_{XS} \right)^\dag \mK_{XS}\mY\\
		&=\left(\mK_{XS}\mK_{XS}+\lambda \mK_{XS} +\zeta\mK_{XS} \mA\mK_{XS} \right)^\dag \mK_{XS}\mY.
	\end{align*}

\section{ADDITIONAL EXPERIMENT RESULTS}
    \label{app::add_exp}
    \subsection{Supplementary results for Section \ref{sec::exp}}
\label{app::supp_exp}

\begin{figure}[H]
	\centering
	\includegraphics[width=\linewidth]{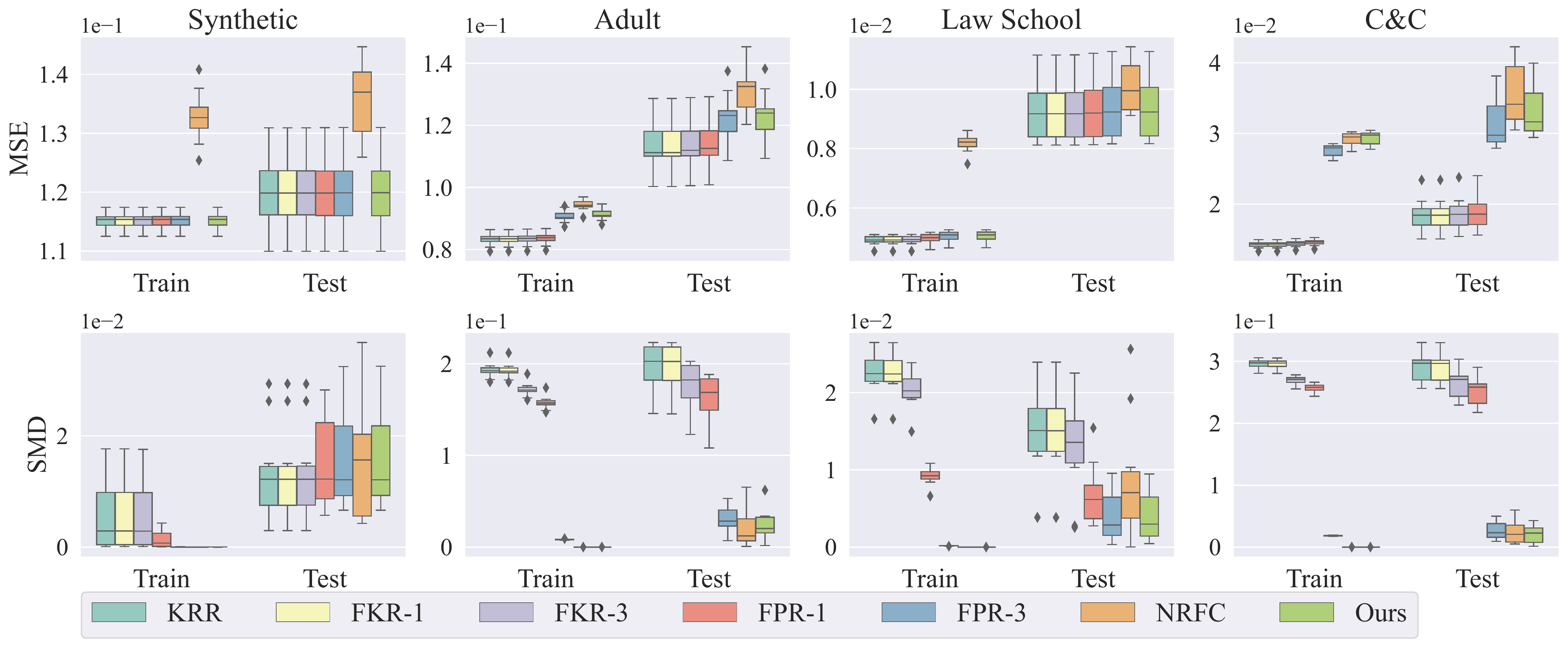}
	\caption{Results of kernel regression for all datasets with one binary sensitive attribute. Figures in the first row show the MSE of different methods whereas the figures in the second row show the SMD of different methods. The methods FKR-1 and FKR-3 stand for FKR method with regularizer coefficients 10 and 1000 respectively. And FPR-1 and FPR-3 stand for FPR method with regularizer coefficients 10 and 1000 respectively. }
	\label{fig::Ker_all}
\end{figure}

\begin{figure}[H]
	\centering
	\includegraphics[width=\linewidth]{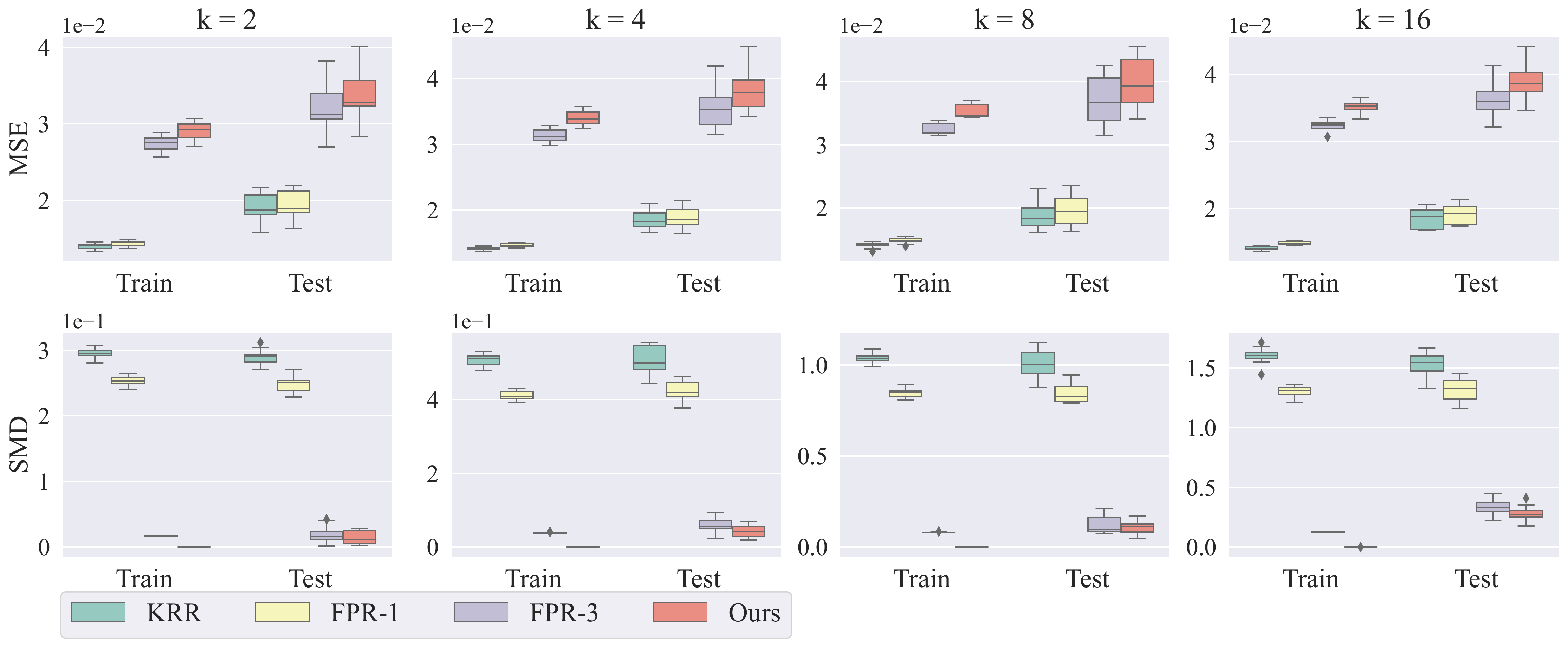}
	\caption{Results of kernel regression on the Communities \& Crime dataset with multiple binary sensitive attributes. Figures in the first row show the MSE of different methods whereas the figures in the second row show the SMD of different methods.}
	\label{fig::multi_ker}
\end{figure}

\begin{figure}[H]
	\centering
	\includegraphics[width=\linewidth]{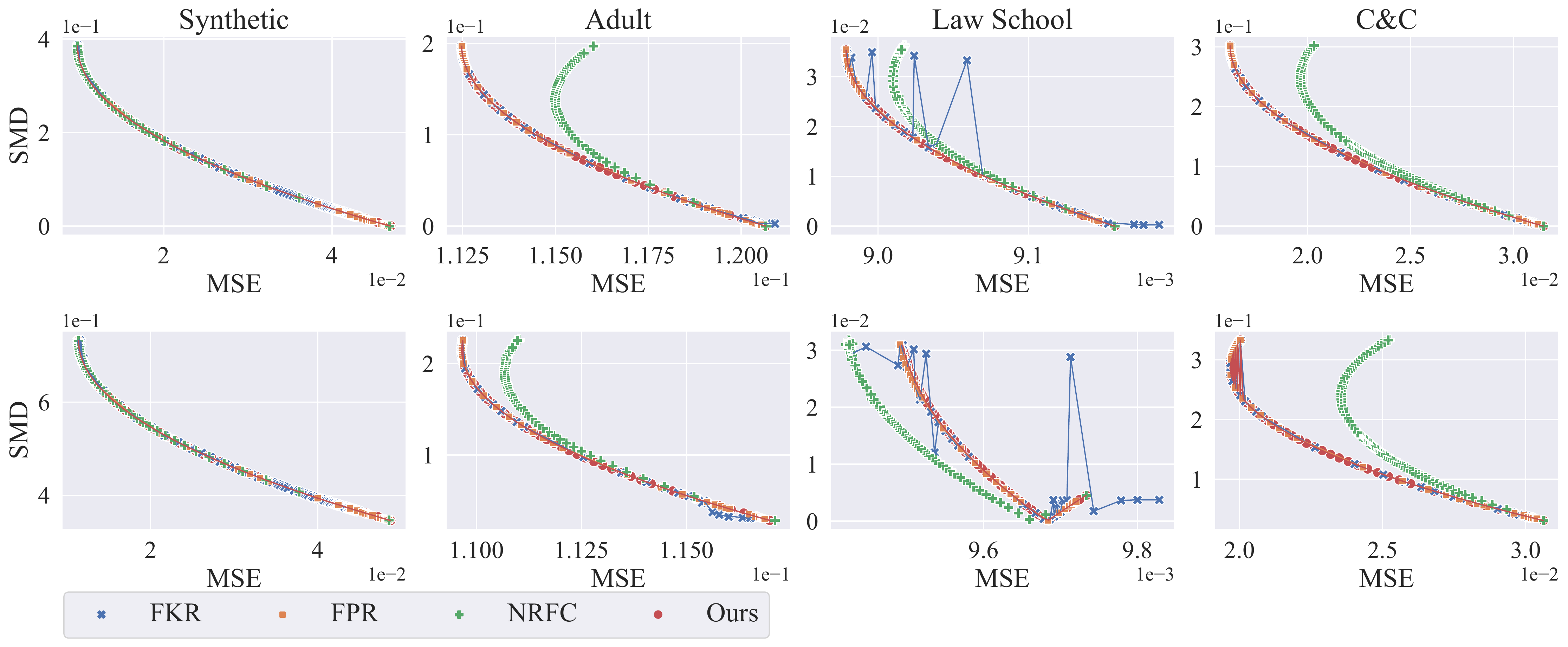}
	\caption{Results of the tradeoff between fairness and accuracy. Figures in the first row show the experiment results on train data whereas the figures in the second row show the tradeoff in test data. In order to compare the results on different datasets, some results in Section \ref{sec::exp} are repeated. We remark that both FKR and NRFC suffer from numerical instability with respect to MSE when fairness constraint is removed or 
    strictly imposed which can be seen from experiment results on the Law School dataset.}
	\label{fig::trade_all}
\end{figure}

In this section, we provide the supplementary experiment results for Section \ref{sec::exp}. In Figure \ref{fig::Ker_all}, we compare different baselines in the kernel regression setting for single binary sensitive attribute. Figure \ref{fig::multi_ker} describes the performance of KRR, FPR and the proposed method in the setting of kernel regression for multiple sensitive attributes. In Figure \ref{fig::trade_all}, we summarize the experiment results for the fairness-accuracy tradeoff on different datasets.

\subsection{Experiments on constant baselines}
In this section, we evaluate the baseline with constant prediction equal to the mean of the labels. Since the "Constant Prediction" baseline can achieve perfect MP fairness in both train data and test data, we only compare the MSE of the "Constant Prediction" baseline and our method for simplicity. The experiment results are summarized in Table \ref{tab::clinear} and Table \ref{tab::ckernel}.

The experiment results show that our method significantly outperforms the "Constant Prediction" baseline in all settings, as expected.

We remark that for linear regression on the Synthetic dataset, the MSE of the "Constant Prediction" is about $77\times$ higher than the MSE of our method since the MSE of "Constant Prediction" baseline is highly dependent on the scale of response.

\begin{table}[H]
\caption{Experiment results on constant baselines for linear regression.}
\label{tab::clinear}
\begin{tabular}{|l|l|c|c|c|c|}
\hline
\multicolumn{1}{|c|}{{ \textbf{Method}}} & \multicolumn{1}{c|}{{ \textbf{Metric}}} & { \textbf{Adult}}                    & { \textbf{Law School}}               & { \textbf{Communities \& Crime}}     & { \textbf{Synthetic}}                \\ \hline
{ Constant}                              & { MSE (Train)}                          & { 0.1858 $\pm$ 0.0016} & { 0.0101 $\pm$ 0.0002} & { 0.0544 $\pm$ 0.0010} & { 4.5342 $\pm$ 0.0450} \\ \hline
{ Constant}                              & { MSE (Test)}                           & { 0.1839 $\pm$ 0.0062} & { 0.0103 $\pm$ 0.0008} & { 0.0536 $\pm$ 0.0039} & { 4.6285 $\pm$ 0.1806} \\ \hline
{ Ours}                                  & { MSE (Train)}                          & { 0.1175 $\pm$ 0.0018} & { 0.0092 $\pm$ 0.0002} & { 0.0313 $\pm$ 0.0009} & { 0.0585 $\pm$ 0.0081} \\ \hline
{ Ours}                                  & { MSE (Test)}                           & { 0.1327 $\pm$ 0.0072} & { 0.0095 $\pm$ 0.0008} & { 0.0344 $\pm$ 0.0026} & { 0.0577 $\pm$ 0.0081} \\ \hline
\end{tabular}
\end{table}
\begin{table}[ht]
\caption{Experiment results on constant baselines for kernel regression.}
\label{tab::ckernel}
\begin{tabular}{|l|l|c|c|c|c|}
\hline
\multicolumn{1}{|c|}{{ \textbf{Method}}} & \multicolumn{1}{c|}{{ \textbf{Metric}}} & { \textbf{Adult}}                    & { \textbf{Law School}}               & { \textbf{Communities \& Crime}}     & { \textbf{Synthetic}}                \\ \hline
{ Constant}                              & { MSE (Train)}                          & { 0.1858 $\pm$ 0.0016} & { 0.0101 $\pm$ 0.0002} & { 0.0544 $\pm$ 0.0010} & { 0.1474 $\pm$ 0.0015} \\ \hline
{ Constant}                              & { MSE (Test)}                           & { 0.1839 $\pm$ 0.0062} & { 0.0103 $\pm$ 0.0008} & { 0.0536 $\pm$ 0.0039} & { 0.1476 $\pm$ 0.0061} \\ \hline
{ Ours}                                  & { MSE (Train)}                          & { 0.0913 $\pm$ 0.0020} & { 0.0050 $\pm$ 0.0002} & { 0.0294 $\pm$ 0.0010} & { 0.1151 $\pm$ 0.0014} \\ \hline
{ Ours}                                  & { MSE (Test)}                           & { 0.1232 $\pm$ 0.0076} & { 0.0093 $\pm$ 0.0010} & { 0.0332 $\pm$ 0.0034} & { 0.1202 $\pm$ 0.0060} \\ \hline
\end{tabular}
\end{table}

\subsection{Experiments on CB fair regression}
\label{exp::CB}
In this section, we show the experimental results of applying our method to CB fairness. The datasets and experiment settings are the same as in Section \ref{sec::exp} except for the choice of $\kappa_S$ for FPR in the proposed method. In this experiment, we choose $\kappa_S$ under Assumption \ref{ass::2}. We compare the proposed method with baselines in terms of MSE and the Norm of the covariance matrix, i.e., 
$$\textbf{Norm of Cov}=\|\Cov(\phi_S(S), g(X, S))\|_{\gH_S}.$$

\begin{figure}[H]
	\centering
	\includegraphics[width=\linewidth]{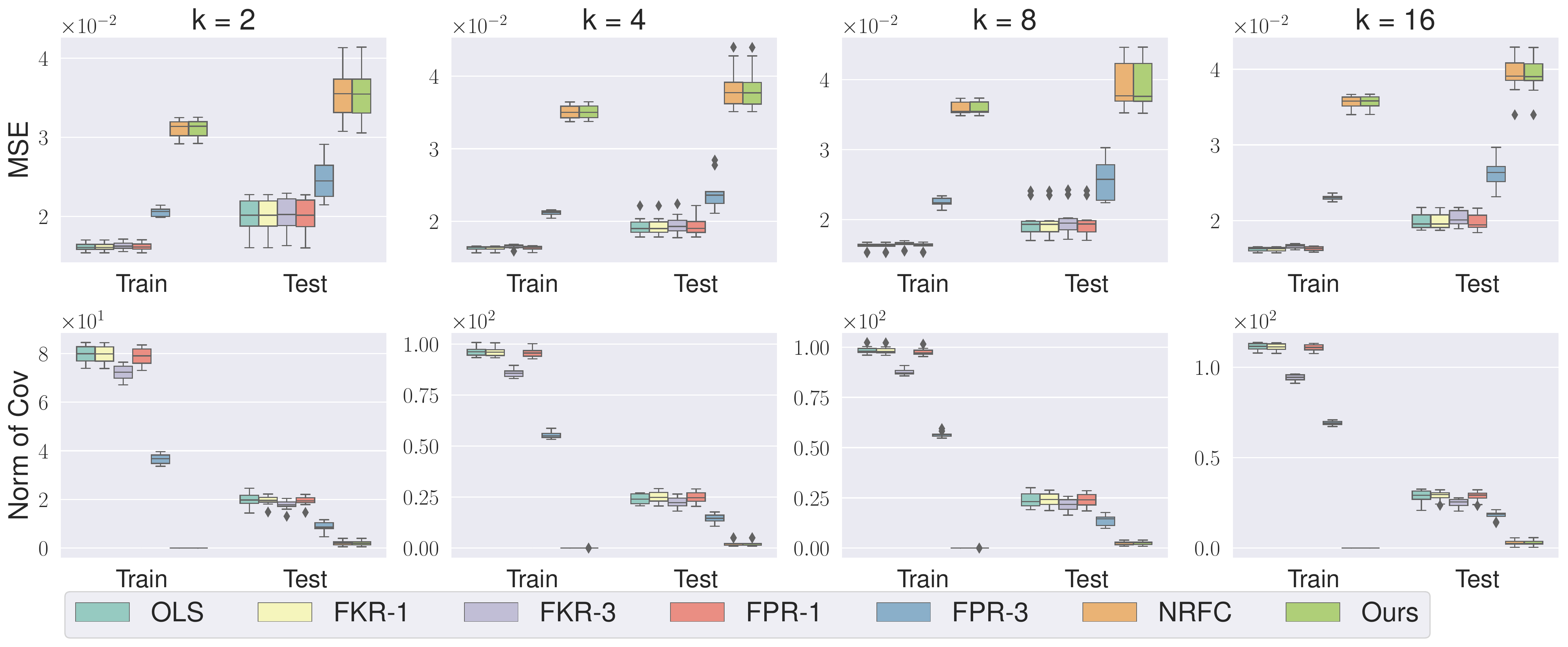}
	\caption{Results of linear regression on the Communities \& Crime dataset with multiple binary sensitive attributes. Figures in the first row show the MSE of different methods whereas the figures in the second row show the \textbf{Norm of Cov} of different methods.}
	\label{fig::multi_linear_cov}
\end{figure}

\begin{figure}[H]
	\centering
	\includegraphics[width=\linewidth]{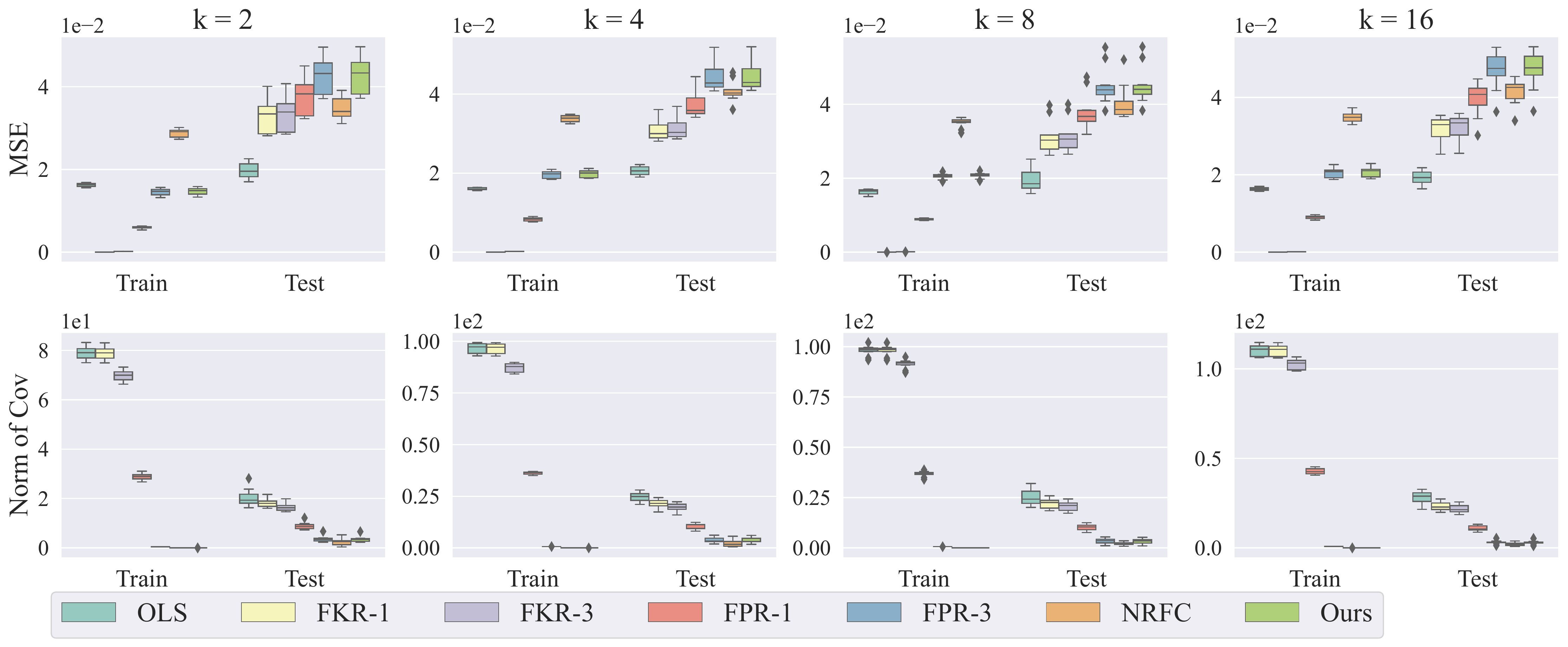}
	\caption{Results of kernel regression on the Communities \& Crime dataset with multiple binary sensitive attributes. Figures in the first row show the MSE of different methods whereas the figures in the second row show the \textbf{Norm of Cov} of different methods.}
	\label{fig::multi_ker_cov}
\end{figure}

Figure \ref{fig::multi_linear_cov} describes the results for the linear regression case which shows that our method achieves almost the same performance as NRFC. In Figure \ref{fig::multi_ker_cov}, we can find that the MSE of our methods is much lower than the MSE of NRFC in the train data. However, our method receives higher MSE than NRFC in the test data, which shows an overfitting problem in this setting. A similar trend can be found with respect to the norm of covariance.

\subsection{Experiments on DP fairness regression}
    \label{app::DP_exp}
\begin{figure}[H]
	\centering
	\includegraphics[width=\linewidth]{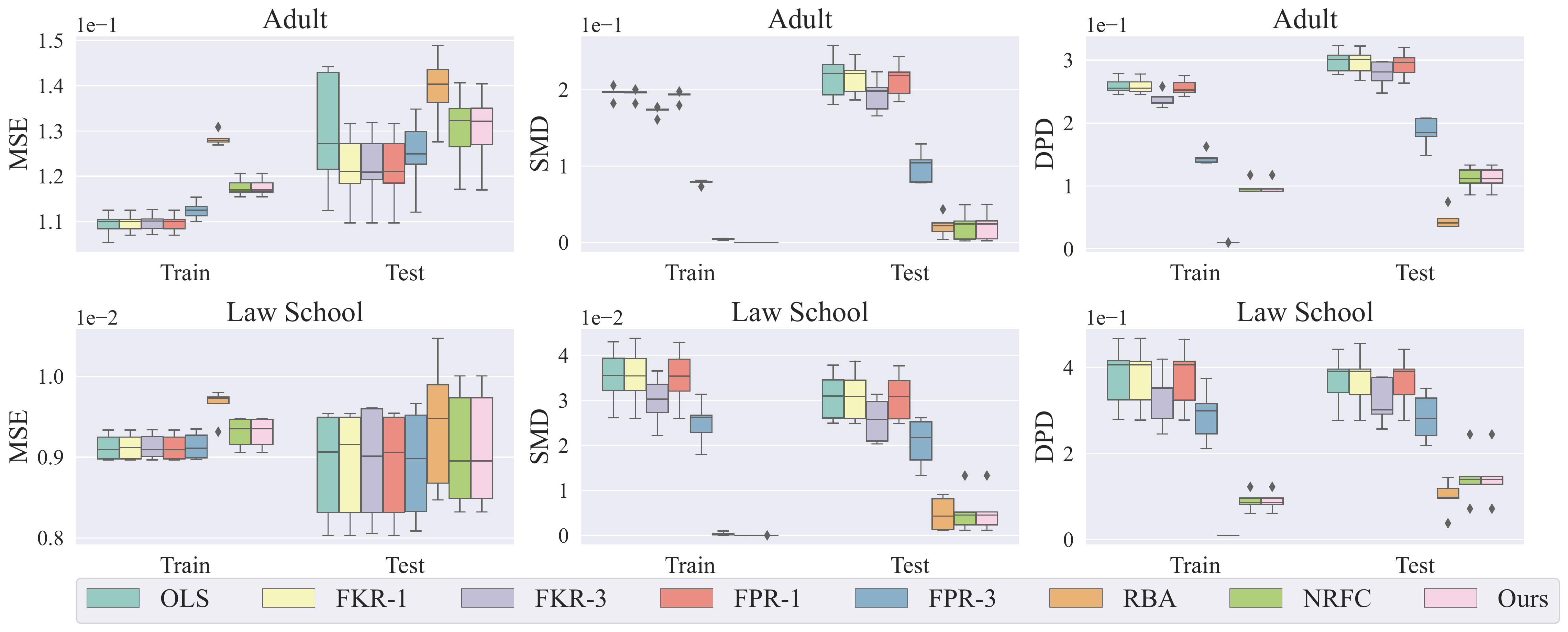}
	\vskip -0.05in
	\caption{Results under the setting of linear regression with a single binary sensitive attribute.}
	\label{fig::LS_binary_all_DP}
	\vskip -0.06in
    \end{figure}

    We also compare the performance of our method with a recent (in-processing) method for DP fairness, i.e., the reduction-based algorithm  (RBA, \cite{agarwal2018reductions}). Note that RBA is designed to produce a DP-fair randomized predictor rather than a simple linear/kernel regression function. We test RBA under the setting of the linear regression with a single binary sensitive attribute, and the experiment results on two benchmark datasets are shown in Figure \ref{fig::LS_binary_all_DP}.     
    
    In this experiment, we found that enforcing DP fairness helps to improve MP fairness and vice versa.  However, as DP is a stronger notion of fairness, a DP-fair regression function has a significantly larger cost of fairness, i.e., a larger loss. Note that all algorithms suffer from distribution shifts in the test data, so both MSE, SMD, and DPD are higher in the testing phase. However, since DP fairness is stronger than MP fairness, RAB can achieve comparable and even lower SMD on the test dataset sometimes, even if our algorithm can eliminate MP unfairness in the train data. This motivates us to investigate the generalization problem for fair algorithms in our future work. We remark that our method is almost $200\times$ faster than RBA in the above experiment.

\subsection{Experiments on other loss functions}
\begin{figure}[H]
	\centering
	\includegraphics[width=\linewidth]{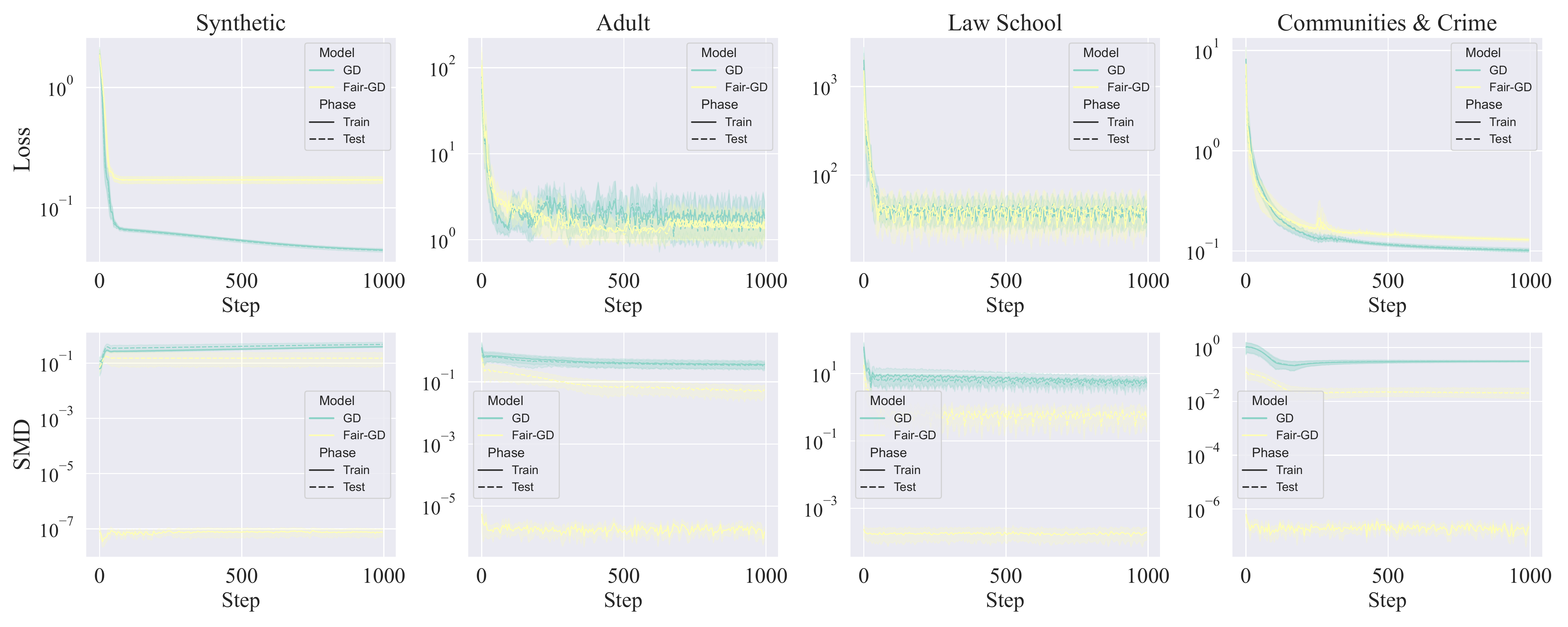}
	\caption{Results of Fair-GD with single binary sensitive attribute using Smooth L1 Loss ($\beta=0.1$). Figures in the first row show the loss of different methods whereas the figures in the second row show the SMD of different methods.}
	\label{fig::linear_pgd}
    \end{figure}

\begin{figure}[H]
	\centering
	\includegraphics[width=\linewidth]{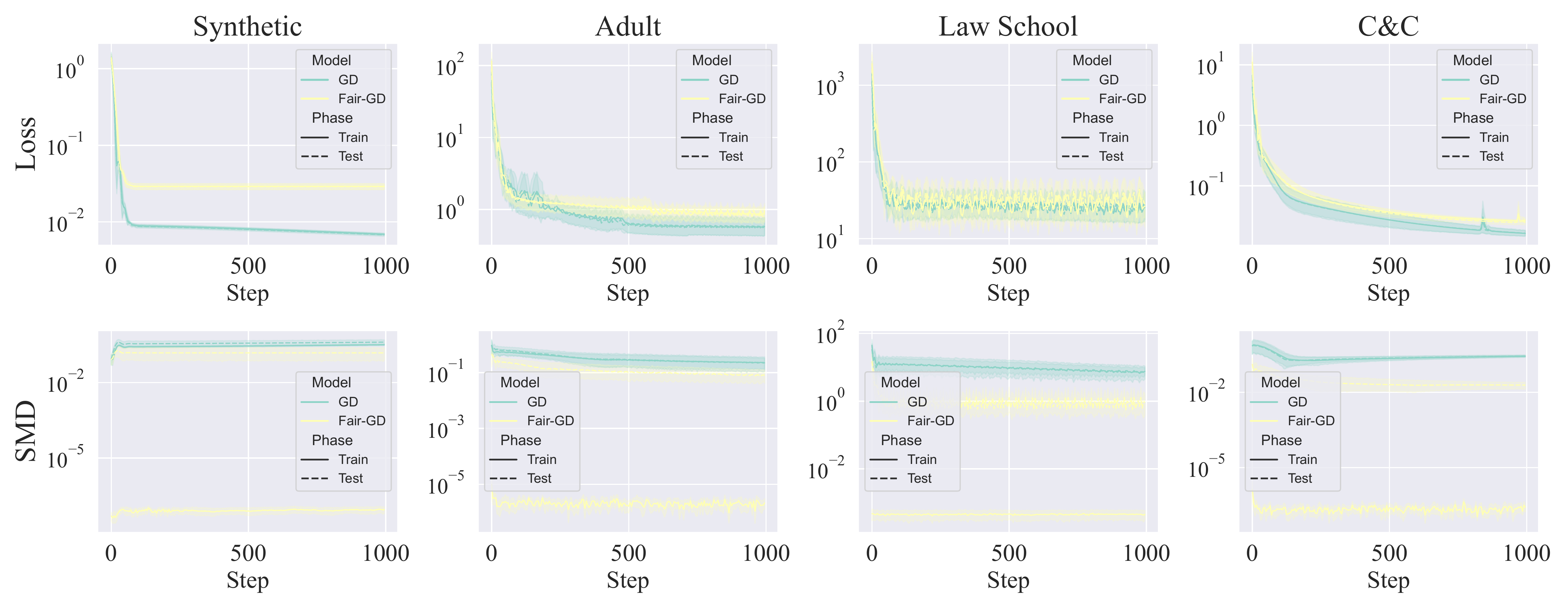}
	\caption{Results of Fair-GD with single binary sensitive attribute using Smooth L1 Loss ($\beta=1$). Figures in the first row show the loss of different methods whereas the figures in the second row show the SMD of different methods.}
	\label{fig::linear_pgd_L1}
    \end{figure}

    In this section, we evaluate the proposed method on other loss functions using gradient descent (Fair-GD). Specifically, we set the loss function to be Smooth L1 Loss, a commonly used loss function that is less sensitive to outliers than the MSE as it treats error as square only inside an interval. 
     We evaluate Fair-GD in the setting of linear regression with single binary sensitive attribute and compare Fair-GD with the gradient descent (GD) algorithm to show its effect on enforcing fairness. In this experiment, we use Adam \citep{kingma2014adam} as our optimizer with a learning rate $1\times 10^{-4}$. The results are summarized in Figure \ref{fig::linear_pgd} and Figure \ref{fig::linear_pgd_L1}, from which we can see that Fair-GD enjoys the same convergence rate as GD while consistently enforcing the fairness constraint.

\subsection{Visualization of distribution}

\begin{figure}[H]
	\centering
	\includegraphics[width=\linewidth]{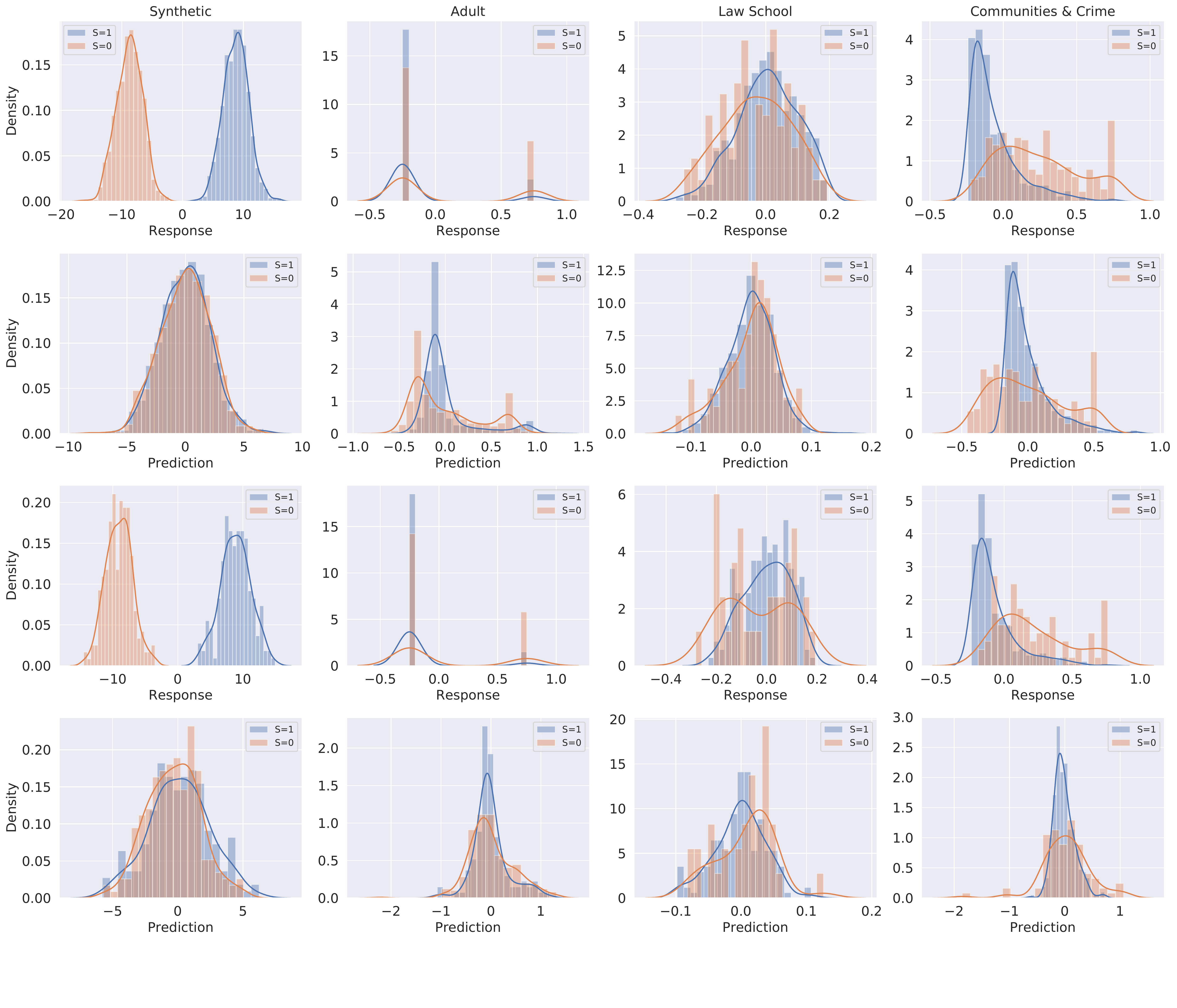}
	\caption{Visualization of response and predicted MP-fair response. Figures in the first row and the third row show the conditional distribution of response variables in the training dataset and test dataset, respectively. Similarly, the figures in the second row and the fourth row show the conditional distribution of the predicted response variables in the training dataset and test dataset, respectively.}
	\label{fig::dist}
    \end{figure}

    In this section, we provide the visualization of MP-fair response for all datasets as an extension to Figure \ref{fig::dist_0}. The results are summarized in Figure \ref{fig::dist}.
    
\subsection{Removing sensitive attributes}
    In this section, we consider the case of  removing the sensitive attributes from the regression function which is a good choice for mitigating unfairness. We remark that such a setting can be regarded as a special case of our general setting. By choosing a kernel $\kappa_{XS}$ which ignores  the input $S$, i.e., $\kappa_{XS}(X,S)=\kappa_{X}(X)$, the proposed method can be adapted  to fair regression without sensitive attributes. We evaluate the proposed method for regression without inputting sensitive attributes on the linear regression with binary sensitive attribute case and summarize the experiment results in Table~\ref{tab::1}. Note that we omit the SMD in  the training  dataset since it is zero in our experiments. The experiment results show that \emph{including the sensitive attribute in regression can help to reduce the MSE while removing the sensitive attribute may help to improve the testing fairness.} 

    \begin{table}[ht]
\renewcommand\arraystretch{1.2}
\resizebox{\textwidth}{!}{%
\begin{tabular}{l|ll|llll}
\hline
           & \multicolumn{2}{c|}{Train}                                        & \multicolumn{4}{c}{Test}                                                                                                            \\
           & \multicolumn{1}{c}{MSE w/ $S$} & \multicolumn{1}{c|}{MSE w/o $S$} & \multicolumn{1}{c}{MSE w/ $S$} & \multicolumn{1}{c}{MSE w/ $S$} & \multicolumn{1}{c}{SMD w/o $S$} & \multicolumn{1}{c}{SMD w/o $S$} \\ \hline
Synthetic  & \textbf{0.0584$\pm$0.0081}     & 2.2994$\pm$0.6630                & \textbf{0.0577$\pm$0.0081}     & 2.4286$\pm$0.7560              & \textbf{0.1508$\pm$0.1348}      & 0.1957$\pm$0.1235               \\
Adult      & \textbf{0.1176$\pm$0.0018}     & 0.1203$\pm$0.0022                & \textbf{0.1327$\pm$0.0072}     & 0.1350$\pm$0.0076              & \textbf{0.0300$\pm$0.0243}      & 0.0355$\pm$0.0221               \\
Law School & \textbf{0.0092$\pm$0.0002}     & 0.0094$\pm$0.0002                & \textbf{0.0095$\pm$0.0008}     & 0.0097$\pm$0.0008              & 0.0055$\pm$0.0034               & \textbf{0.0042$\pm$0.0033}      \\
C\&C       & \textbf{0.0313$\pm$0.0009}     & 0.0375$\pm$0.0010                & \textbf{0.0344$\pm$0.0026}     & 0.0402$\pm$0.0028              & 0.0232$\pm$0.0164               & \textbf{0.0187$\pm$0.0087}      \\ \hline
\end{tabular}}

\caption{Experiments on Mean Parity fair linear regression with (w/) and without (w/o) the sensitive attitude $S$. }
\label{tab::1}
\end{table}

\section{TABLE OF NOTATIONS}
    \label{app::sec0}
    We summarize the notations used throughout this paper in the following table.
\begin{longtable}{|l|l|l|}
\caption{Table of notations}\\
\hline\\
Notation                                & Description/Definition \\
$\Delta$                                & An arbitrary function in $\gG_{MP}$\\
$\Sigma $                               & The covariance operator\\
$\hat{\Sigma} $                         & The empirical estimation of $\Sigma$\\
$\tilde{\Sigma }  $                     & The uncentralized covariance operator\\
$\Phi$                                  & The feature matrix of the data     \\
$\Omega$                                & A set from which a random variable is chosen \\
$\gF$                                   & Borel $\sigma$-filed on $\Omega$\\
$\E$                                    & The expectation function\\
$\sP$                                   & The probability function\\
$\sR$                                   & Set of real numbers\\
$\gG_{MP} $                             & A MP-fair space\\
$A $                                    & A linear operator\\
$F$                                     & The  cumulative distribution function of random variable\\
$J $                                    & The generalized objective function for regression problem\\
$L $                                    & The mean square loss function\\
$P$                                     & The projection operator\\
$\hat{P} $                              & The empirical estimation of $P$\\
$S$                                     & Random variable for sensitive attributes\\
$X$                                     & Random variable for non-sensitive attributes\\
$Y$                                     & Random variable for label/response\\
$\mH$                                   & $\mH=\mI_{n\times n}-\frac{1}{n}\vone_{n\times n}$\\
$\mK$                                   & The Gram matrix\\
$\mP$                                   & $\mP=(\mI_{n\times n}-\sum_{j=1}^{m}\va_m\va_m^T\mK_{XS})$\\
$\mY$                                   & The vector of response in dataset\\
$\hat{\mY}$                             & The predicted value of $\mY$\\
$\text{DPD} $                           & The DP disparity\\
$\text{MPD} $                           & The  MP disparity\\
$\alpha $                               & A scalar in $[0,1]$ to control the accuracy-fairness tradeoff\\
$\beta $                                & A real number\\
$\eps $                                 & Random noise\\
$\zeta$                                 & The parameter for fairness penalty term\\
$\eta $                                 & A real number\\
$\theta $                               & An (normalized) eigenfunction of $\Sigma_{(XS)S}A\Sigma_{S(XS)}$\\
$\kappa$                                & Kernel function\\
$\lambda$                               & The regularization coefficient\\
$\mu $                                  & A kernel mean embedding\\
$\phi$                                  & Feature map \\
$\psi$                                  & Eigenvalue \\
$\bar{\phi}$                            & The empirically centralized feature map \\
$\va$                                   & The weight vector for an (normalized) eigenfunction with respect to $\Phi_{XS}$\\
$\bar{\va}    $                         & The weight vector for an (normalized) eigenfunction with respect to $\bar{\Phi}_{XS}$\\
$g,f $                                  & Functions\\
$g^* $                                  & The least-squares regression function\\
$g^*_{\gG}$                             & An optimal solution for Problem \ref{eq::obj}\\
$\hat{g}^*_{\gG}$                       & The empirical estimation of $g^*_{\gG}$\\
$g^\alpha $                             & An optimal solution to Problem \ref{eq::trade0} \\
$g_{MP} $                               & The projection of $g$ on $\gG_{MP}$\\
$g_{MP^\perp} $                         & The projection of $g$ on $\gG_{MP^\perp}$\\
$h$                                     & A function defined as $h=\E(\phi_{XS}(X, S)Y)$\\
$k $                                    & The cardinality of $\Omega_{S}$\\
$m $                                    & The rank of $\Sigma_{(XS)S}$\\
$n $                                    & The number of training samples\\
$s$                                     & A realization  of $S$\\
$x$                                     & A realization  of $X$\\
$y$                                     & A realization  of $Y$\\
$\vw $                                  & A weight vector\\
$\ell $                                 & A differentiable loss function\\
$\ker $                                 & The kernel of a linear operator\\
$\ran $                                 & The range of a linear operator\\
$\otimes  $                             & The outer product\\
$\langle \cdot, \cdot \rangle $         & The inner product\\
$0_{\gH} $                              & The zero function in $\gH$\\
$\dagger $                              & The Moore-Penrose Inverse of an operator\\
$\mathbb{I}{(\cdot)}$                  & The indicator function\\
$\perp$                                 & The orthogonal complement of a space\\
$\ind $                                 & Independence between two random variables\\
\hline
\end{longtable}

\end{document}